\documentclass[11pt]{article}
\usepackage[]{acl}
\usepackage{times}
\usepackage{textcomp}
\usepackage{latexsym}
\usepackage{hyperref}
\usepackage[T1]{fontenc}
\usepackage[utf8]{inputenc}
\usepackage{microtype}
\usepackage{inconsolata}
\usepackage{graphicx}
\usepackage{soul}
\usepackage{float}
\usepackage{tabularx}
\usepackage{multirow}
\usepackage{scalefnt}
\usepackage{subfigure}
\usepackage{booktabs}
\usepackage{graphicx}  
\usepackage{adjustbox} 
\usepackage{amsmath}    
\usepackage{amssymb}    
\usepackage{xcolor}
\usepackage{tcolorbox}
\tcbuselibrary{skins,breakable}

\title{Beyond Topical Similarity: Contrastive Evidence Retrieval with Interpretable Attention Alignment in RAG} 

\author{Francielle Vargas\thanks{Corresponding author: francielle.vargas@uchile.cl}\\ University of Chile \\
\And
João Robiatti \\ São Paulo State University\\
\And
Diego Alves  \\ Saarland University \\
\AND
Lucas Pascotti Valem \\ University of São Paulo \\
\And
Maximilian Seeth \\ University of Munich \\
\And 
Sebastián Ferrada \\ University of Chile\\
\AND
Ameeta Agrawal \\ Portland State University \\
\And
Daniel Pedronette \\ São Paulo State University\
\And
André Freitas \\ Idiap Research Institute\\
}

\begin{document}
\maketitle
\begin{abstract}
Ensuring factuality and interpretability in RAG remains an open and urgent problem. We introduce Contrastive Evidence Rationale Attention (CERA)\footnote{Dataset, code, and models are publicly available at \url{https://github.com/franciellevargas/CERA}.}, the first retrieval framework to employ subjectivity-based hard negative selection and inject an evidential inductive bias into contrastive learning through an auxiliary attention alignment loss. CERA fine-tunes a dense retriever using two training objectives: triplet-based contrastive learning and interpretable attention alignment, which supervises CLS-to-token attention using a part-of-speech-weighted masking distribution over human-annotated factual rationales as evidence signals. Experiments on a large corpus of clinical trial reports demonstrate that the subjectivity-based hard negative selection substantially improves retrieval effectiveness compared to both Contriever and hard negative selection baselines. Furthermore, rationale alignment improves faithfulness while maintaining competitive retrieval performance, supporting the hypothesis that attention can serve as a more faithful explanation of model behavior when guided by human rationales. Moving beyond topical similarity, CERA enables the retriever to identify the specific tokens that constitute supporting evidence, promoting more interpretable evidence selection in RAG systems.

\end{abstract}


\section{Introduction}

Large language models (LLMs) augmented with retrieval have recently emerged as a promising solution to improve factuality \cite{krishna-etal-2025-fact,ranaldi-etal-2025-eliciting} and reduce hallucinations \cite{niu-etal-2024-ragtruth,hu-etal-2025-removal}. RAG improves LLMs by constraining their outputs in external documents, offering evidence through retrieved text passages rather than depending only on the model’s parametric memory \cite{lewis2020retrieval,ram-etal-2023-context}.

\begin{figure}[!htb]
\begin{center}
	\includegraphics[height=5.5cm]{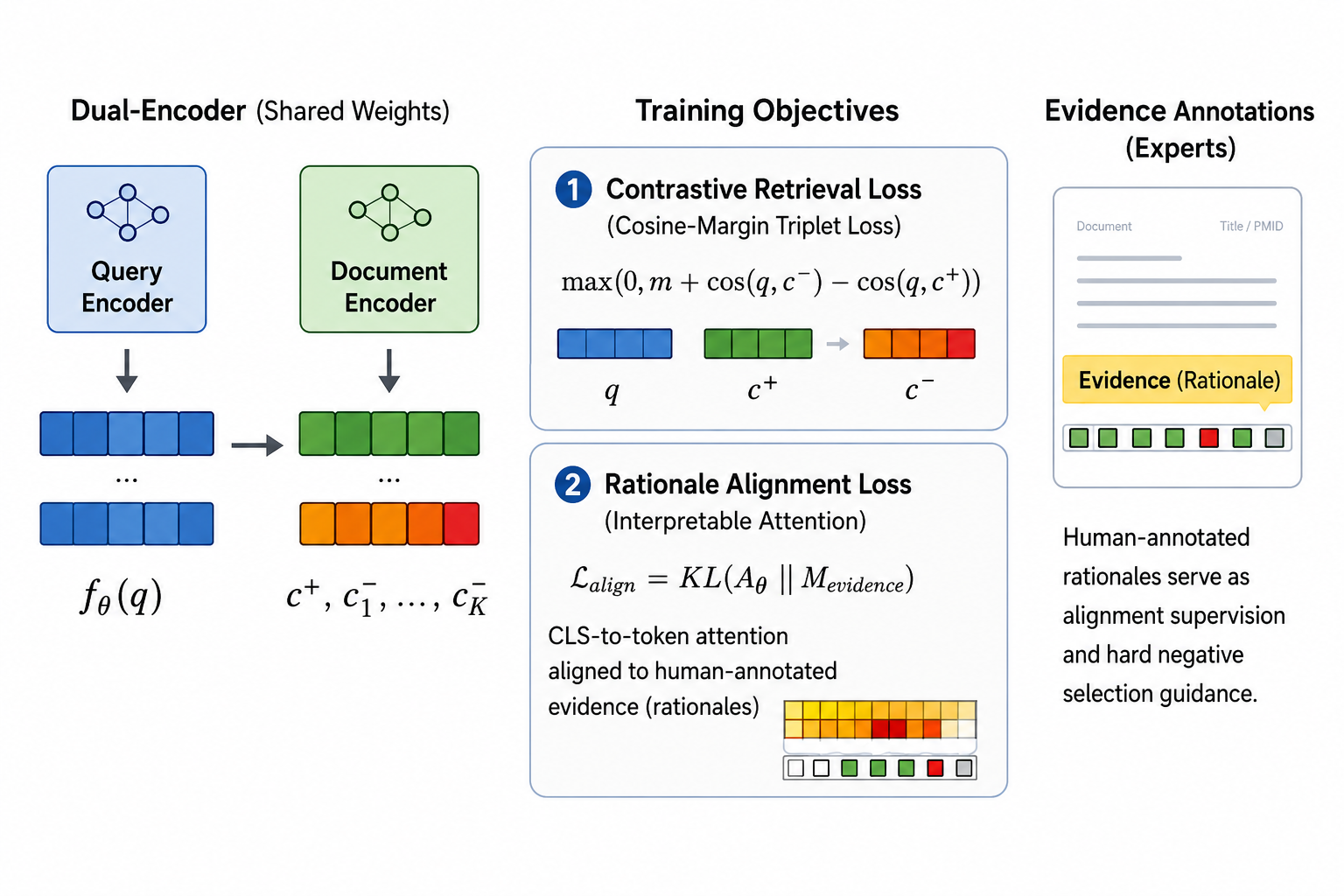}
\caption{Contrastive Evidence Rationale Attention (CERA) is the first retrieval framework that employs subjectivity-based hard negative selection and inject an evidential inductive bias into contrastive learning through an auxiliary attention alignment loss.} 
\label{fig:CERA_intro}
\end{center}
\end{figure} 

In this regard, dense textual retrieval is central to RAG systems \cite{lewis2020retrieval,ram-etal-2023-context}, enabling LLMs to incorporate external knowledge without updating model parameters. However, recent studies show that the effectiveness of RAG is often compromised by noisy retrieval, where genuinely evidential passages are mixed with content that is merely topically related, contradictory, or factually unreliable \cite{liu-etal-2023-evaluating,petroni2020contextaffectslanguagemodels}. Most RAG systems are not explicitly parameterized to distinguish evidential adequacy from semantic proximity, leading to systematic confounding between topical relevance and factual validity. For example, given the query ``Is lemon effective for curing cancer?'', a standard retriever may prioritize passages containing lexical co-occurrences or misinformation related to lemons and cancer rather than clinically substantiated evidence or medical consensus. Thus, generation models conditioned on retrieval distributions lacking evidential precision are more likely to produce incomplete, biased, or spurious content, amplifying hallucinations and compromising factual reliability \cite{hu-etal-2025-removal,abolghasemi-etal-2025-evaluation}. This limitation is particularly critical in high-stakes domains such as healthcare and governance \citep{dahl2024large}, where unsupported or weakly grounded outputs may compromise safety and lead to significant ethical, societal, and legal consequences \cite{sun-etal-2025-fact,delbrouck-etal-2022-improving,shi2023,Legislation2016,chen-etal-2025-llms,cheng-amiri-2025-equalizeir,ghate-etal-2025-biases,kim-etal-2024-discovering}.

Recent approaches have addressed these limitations through contrastive learning \cite{gu-etal-2025-toward,sriram-etal-2024-contrastive,izacard2022unsuperviseddenseinformationretrieval}, typically using Transformer-based architectures fine-tuned with objectives that train retrievers to distinguish relevant from irrelevant passages by maximizing the separation between positive and negative examples \cite{wu-etal-2025-medical,sun-etal-2025-fact,sriram-etal-2024-contrastive}. One challenging aspect of fine-tuning embedding models lies in identifying high-quality hard negatives to support effective contrastive learning \cite{10.1145/3746252.3761254}. In practice, these methods often treat all non-relevant documents as equally negative, overlooking the diverse semantic relations they may hold to a target claim: some may offer verifiable supporting evidence, while others may contradict the claim with opposing findings. This coarse treatment limits the model’s ability to capture nuanced evidential structure. The most relevant hard negative mining strategies are based on relevance scores for the query \cite{10.1145/3746252.3761254}, or on selecting negative passages based on their similarity to the positive passage rather than to the query itself \cite{nguyen-etal-2023-passage}. Nevertheless, these approaches remain limited, as the absence of semantically informed criteria that explicitly distinguish between neutral, supportive, and contradictory evidence may lead dense retrieval systems to learn spurious semantic relations that conflate factual disagreement with irrelevance, ultimately undermining both robustness and factual consistency in downstream tasks

Beyond the challenges of hard negative selection, model interpretability remains a critical concern in retrieval-based approaches~\cite{xin-etal-2025-sparse,li-etal-2025-iris,kim-lee-2024-rag}. Recent advances in retrieval-augmented models primarily focus on optimizing \textit{what} evidence is retrieved, typically measured by downstream performance gains. However, these approaches largely overlook \textit{why} specific evidence is selected and \textit{how} the retrieved content substantively supports the model’s decision~\cite{deng-etal-2024-learning}. Moreover, existing interpretability techniques~\cite{10.1145/2939672.2939778,lundberg2017,kim-etal-2024-discovering} rely predominantly on post-hoc explanations, highlighting influential words or sentences without providing intrinsic interpretability mechanisms for self-explaining retrieval models. This gap underscores the need to move beyond optimizing \textit{what} evidence is retrieved toward explicitly modeling \textit{why} it is selected and \textit{how} it supports model decisions, rather than treating retrieval as a black-box process.




 To address these limitations, we propose \textbf{Contrastive Evidence Rationale Attention (CERA), the first interpretable dense retrieval framework that jointly optimizes contrastive retrieval and evidence-aware attention alignment in RAG systems}. Although prior work presents divergent perspectives on whether attention constitutes a faithful interpretability mechanism \cite{wiegreffe-pinter-2019-attention}, recent studies suggest that attention can provide a degree of interpretability under certain conditions \cite{serrano-smith-2019-attention,chrysostomou-aletras-2021-improving,Vashishth2019AttentionIA}. Nevertheless, attention distributions in Transformer-based models are often poorly aligned with human reasoning processes \cite{herrewijnen-etal-2026-bert,vargas-etal-2026-self-explaining,Eilertsen_Bjrgfinsdttir_Vargas_Ramezani-Kebrya_2026}. Rather than assuming attention is inherently interpretable, our framework explicitly promotes interpretability through an evidential inductive bias grounded in human-annotated evidential rationales. This design encourages attention distributions to align with human explanatory evidence during training. Specifically, CERA is built upon two core concepts introduced in this paper: \textit{Interpretable Attention Alignment}, defined as the explicit supervision of neural attention mechanisms through human rationales used as ground-truth explanatory signals; and \textit{Evidential Inductive Bias}, defined as a learning preference that guides the model toward representations grounded in human-annotated evidence rather than relying solely on topical or distributional similarity.

Figure~\ref{fig:CERA_intro} presents the overall architecture of CERA. Given a query-document pair, documents are segmented into chunks and encoded using a dual-encoder architecture trained with triplet-based contrastive learning. Unlike conventional dense retrievers that primarily optimize topical similarity, CERA incorporates both subjectivity-aware hard negative selection and token-level rationale supervision through an auxiliary KL-divergence objective applied over CLS-to-token attention distributions. To construct the rationale supervision signal, we employ a POS-tag-guided weighting scheme over gold evidence rationales, allowing the model to emphasize linguistically informative tokens during attention alignment. Experiments on a large corpus of clinical trial reports demonstrate that CERA consistently improves retrieval effectiveness over both the original Contriever and hard-negative baselines, achieving gains of +0.07681 in Recall@1, +0.10730 in Recall@3, and +0.13070 in Recall@5. Furthermore, rationale alignment improves explanation faithfulness while preserving competitive retrieval performance, reducing Sufficiency scores from 0.2073 to 0.0939 under stronger alignment regularization. Overall, these findings suggest that evidential attention alignment enables retrieval beyond superficial topical similarity and promotes more faithful evidence selection in RAG systems.


Our contributions are: (i) The Contrastive Evidence Rationale Attention (CERA), the first contrastive retrieval framework to combine subjectivity-based hard negative selection with interpretable attention alignment, injecting an evidential inductive bias into dense retrieval using auxiliary rationale supervision. (ii) A rationale-aligned attention objective that supervises CLS-to-token attention using POS-weighted human-annotated factual rationales, enabling the retriever to identify the relevant chunks and the specific tokens that constitute supporting evidence, while improving faithfulness and preserving competitive retrieval performance. (iii) The code and experiments are publicly available to facilitate future research and reproducibility.

\section{Related Work}
\label{sec:relatedwork}

This work investigates Evidence-Based RAG through the lens of contrastive learning, focusing on how dense retrievers can be trained to identify not only topically relevant passages but also the specific evidence supporting model predictions. In this settings, \citet{wu-etal-2025-medical} introduce MedGraphRAG, a graph-based RAG framework for producing evidence-based medical responses. The method combines triple graph construction to link user inputs to verified medical sources and definitions with a hierarchical retrieval strategy that balances contextual coverage and precision. Evaluated on nine medical Q\&A benchmarks and two health fact-checking datasets, MedGraphRAG consistently outperforms standard RAG, GraphRAG, and fine-tuned medical LLMs, achieving state-of-the-art performance while improving factuality, interpretability, and reliability for clinical use. In a complementary multimodal setting, \citet{sun-etal-2025-fact} propose FactMM-RAG, an evidence-based multimodal RAG approach for radiology report generation. The method leverages RadGraph to extract entities and relations and computes factual similarity scores to form aligned image-report pairs, which are used to train a MARVEL-based retriever with contrastive learning and in-batch negatives. When combined with LLaVA for autoregressive generation, FactMM-RAG yields significant gains on MIMIC-CXR and CheXpert, surpassing MedCLIP, CXR-CLIP, and BiomedCLIP in factual accuracy and clinical coherence. Beyond biomedical applications, \citet{chatzikyriakidis-natsina-2025-poetry} explore the use of RAG and contrastive learning for controlled literary generation. Their approach retrieves thematically and stylistically similar—alongside contrastive—poems to guide GPT-4-turbo and GPT-4o in generating Modern Greek interwar poetry. Both expert evaluations and quantitative metrics indicate improved stylistic fidelity and thematic consistency. Finally, focusing on fact verification, \citet{sriram-etal-2024-contrastive} introduce a contrastively trained dense retriever for complex fact-checking. The method extends Contriever with supervision from GPT-4 distillation, LERC-based answer equivalence, and human-annotated AVeriTeC data. Their Contrastive Fact-Checking Reranker improves second-stage retrieval, achieving gains of 6\% in veracity classification accuracy and 9\% in top-document relevance across AVeriTeC, FEVER, ClaimDecomp, and HotpotQA, demonstrating enhanced reasoning and robustness for fact-checking scenarios. Additional related work on \textbf{Interpretable RAG} and \textbf{Hard Negative Selection} is provided in Appendix \ref{sec:appendix_relatedwork}.

\section{Evidence Inference Dataset}

We use the Evidence Inference 2.0 dataset\footnote{\url{https://evidence-inference.ebm-nlp.com/}} \cite{lehman-etal-2019-inferring}, a benchmark for evidence-based inference in clinical trial reports. Each instance consists of an intervention, comparator, outcome (ICO) query, and requires predicting the reported effect significantly increased, significantly decreased, or no significant difference) supported by textual files containing evidence.  Table~\ref{tab:dataset_structure} summarizes the dataset structure and annotation schema, while Table~\ref{tab:example_instance_vertical} presents an example combining an ICO query with a clinical trial document and expert-annotated evidence rationales.

\begin{table}[!htb]
\scalefont{0.68}
\centering
\begin{tabular}{p{1.3cm} p{5.5cm}}
\toprule
\textbf{Component} & \textbf{Description} \\
\midrule
Prompts & Defines ICO tuples\\
Annotations & Expert labels, evidence spans, and validation signals. \\
Split Files & Train/dev/test partitions over prompts. \\
Text Files & Full-text articles linked via PMCID. \\
\bottomrule
\end{tabular}
\caption{Structure of the Evidence Inference 2.0 dataset.}
\label{tab:dataset_structure}
\end{table}


\begin{table}[!htb]
\centering
\footnotesize
\scalefont{0.78}
\setlength{\tabcolsep}{4pt}
\renewcommand{\arraystretch}{1.2}
\begin{tabular}{p{1.4cm} p{5.7cm}}
\toprule
\textbf{Field} & \textbf{Example} \\
\midrule
\textbf{Query} &
Compare the effect of Tar Wars program versus No intervention on Tobacco-related knowledge. \\
\textbf{Outcome} &
Tobacco-related knowledge \\
\textbf{Intervention} &
Tar Wars program \\
\textbf{Comparator} &
No intervention \\
\textbf{Document Excerpt} &
(...) Pooled analyses showed that short-term retention on most items was achieved. Retention of knowledge items, such as recognition of smoking effects and costs, was maintained for four months. \\
\textbf{Evidence Span} &
Retention of knowledge items, such as recognition of smoking effects and costs, was maintained for four months. \\
\textbf{Label} &
Significantly increased \\
\bottomrule
\end{tabular}
\caption{Example instance from the Evidence Inference 2.0 dataset with structured query formulation.}
\label{tab:example_instance_vertical}
\end{table}

\section{Contrastive Evidence Rationale Attention (CERA)}
\label{sec:experiments}
\subsection{Formal Definition}
Let $D$ be a document split into a set of chunks $\mathcal{C} = \{C_1, C_2, \dots, C_n\}$, where each chunk $C_i$ corresponds to a contiguous token span within $D$. Given a query $q$ and a document $D$, the goal is to rank the chunks in $\mathcal{C}$ such that the chunk containing the gold evidence is ranked highest. The gold evidence is defined as a contiguous character span within $D$, denoted by $[s,e] \subset D$, where $s$ and $e$ are the start and end character indices that delimit the exact substring in $D$ that supports the query. For instance, consider the document $D$ = \textit{``Aspirin reduces stroke risk compared to placebo in clinical trials''} and suppose the gold evidence is \textit{``reduces stroke risk compared to placebo''}. In this case, $s$ is the index where “reduces” starts in $D$, and $e$ is the index where “placebo” ends.

\subsection{Chunk Labeling}
In CERA, each chunk $C_i$ is associated with a contiguous span $[s_i, e_i]$ in the document. A chunk is labeled as \textit{positive evidence} if it overlaps with the expert-annotated evidence span $[s,e]$:
\begin{equation}
\scriptsize
[s_i, e_i] \cap [s, e] \neq \emptyset.
\end{equation}

Negative chunks are selected from those not overlapping with $[s,e]$, ranked by a \textit{subjectivity score}\footnote{We used TextBlob (\url{https://textblob.readthedocs.io/en/dev/}) to rank chunks within each PMCID document according to their subjectivity scores, selecting the five highest-scoring subjective passages as hard negatives.}, where higher values mean greater subjectivity, taking the top-$K$ chunks as hard negatives. We use $K=5$ throughout the paper.

\subsection{Dense Retrieval Model}

We employ a dual-encoder architecture where $f_\theta(\cdot)$ is a Transformer-based encoder that uses the \texttt{[CLS]} token (rather than mean pooling), as the sentence-level representation:

\begin{equation}
\scriptsize
\mathbf{q} = f_\theta(q), \quad \mathbf{c}_i = f_\theta(C_i),
\end{equation}

The resulting embeddings are L2-normalized:
\begin{equation}
\scriptsize
\mathbf{q}, \mathbf{c}_i \in \mathbb{R}^d, \quad
\|\mathbf{q}\|_2 = \|\mathbf{c}_i\|_2 = 1.
\end{equation}

The relevance score between the query and a chunk is computed via cosine similarity:
\begin{equation}
\scriptsize
\mathrm{sim}(q, C_i) = \mathbf{q}^\top \mathbf{c}_i.
\end{equation}

\subsection{Triplet Retrieval Loss}
For each query-document pair, we construct training triplets $(q, C^+, C^-)$, where $C^+$ is a positive chunk overlapping the expert-annotated evidence span, and $C^-$ is a hard negative chunk selected from non-overlapping chunks within the same document and ranked by a subjectivity score.

The model is trained using the triplet loss:
\begin{equation}
\scriptsize
\mathcal{L}_{\text{triplet}} =
\max \left( 0,
\mathrm{sim}(q, C^-) - \mathrm{sim}(q, C^+) + m
\right),
\end{equation}
where $m$ is a margin hyperparameter.

\subsection{Token-Level Evidence Rationale Masking}
For each positive chunk $C^+ = \{t_1, \dots, t_T\}$, we first define a token-level evidence indicator:
\begin{equation}
\scriptsize
\mathbf{r} \in \{0,1\}^T,
\end{equation}
where $r_j = 1$ if token $t_j$ lies within the gold evidence span $[s,e]$, and $0$ if not. We then incorporate linguistic structure by assigning part-of-speech (POS)\footnote{We used spaCy POS tagging: \url{https://spacy.io/}} dependent weights to each token. We thus define the final weighted rationale signal as:
\begin{equation}
\scriptsize
\hat{r}_j = r_j \cdot w(t_j),
\end{equation}
where $w(t_j)$ denotes the POS-dependent weight of token $t_j$. The weights are defined as follows: NOUN/PROPN/VERB = 1.0, ADJ = 0.9, ADV = 0.8, NUM = 0.7, PART = 0.5, PRON = 0.4, AUX = 0.4, ADP = 0.3, DET = 0.2, CCONJ = 0.2, SCONJ = 0.2, X = 0.5, and PUNCT/SPACE = 0.0.


\subsection{Attention Extraction}
Let $\mathbf{A} \in \mathbb{R}^{H \times T}$ denote the attention weights from the CLS token to the input tokens at the final Transformer layer, averaged across $H$ attention heads:
\begin{equation}
\scriptsize
\mathbf{a} = \frac{1}{H} \sum_{h=1}^H \mathbf{A}_h[\text{CLS}, :].
\end{equation}

We normalize the attention scores using softmax:
\begin{equation}
\scriptsize
\tilde{\mathbf{a}} = \mathrm{softmax}(\mathbf{a}).
\end{equation}

\subsection{Interpretable Attention Alignment and Evidential Inductive Bias}
We introduce \textit{interpretable attention alignment}, a framework that improves the interpretability of neural attention mechanisms by aligning attention distributions with human rationales used as supervision signals. Instead of treating attention as a latent optimization artifact, the model is guided to focus on human-understandable and decision-relevant textual evidence, promoting greater transparency, faithfulness, and trustworthiness in NLP systems. Similar approaches have already been explored by \citet{vargas-etal-2026-self-explaining,Eilertsen_Bjrgfinsdttir_Vargas_Ramezani-Kebrya_2026,herrewijnen-etal-2026-bert}.

We introduce the notion of \textit{evidential inductive bias}: a learning preference that explicitly guides the model toward representations grounded in human-annotated evidence, rather than relying solely on topical or distributional similarity. This reflects the broader principle that systematic generalization depends on inductive bias \cite{mitchell1980bias}.

We operationalize this bias via Kullback–Leibler (KL) divergence \cite{kullback1951information} between the model's attention ($\tilde{\mathbf{a}}$) and the gold evidence rationale masking distribution $\tilde{\mathbf{r}}$:
\begin{equation}
\scriptsize
\mathcal{L}_{\text{alignment}} =
\mathrm{KL}\left(
\tilde{\mathbf{r}} \,\|\, \tilde{\mathbf{a}}
\right),
\end{equation}

\subsection{Multi-Objective Training }
We formulate CERA as a multi-objective training framework with two complementary objectives: (i) contrastive retrieval optimization and (ii) evidential interpretable attention alignment.

The final objective optimizes both components:
\begin{equation}
\scriptsize
\mathcal{L} =
\mathcal{L}_{\text{triplet}} +
\lambda \, \mathcal{L}_{\text{alignment}}
\end{equation}
where $\mathcal{L}_{\text{triplet}}$ optimizes retrieval accuracy by separating positive and negative chunks in the embedding space, and $\mathcal{L}_{\text{alignment}}$ enforces explanation faithfulness via POS-weighted rationale supervision. The hyperparameter $\lambda$ controls the trade-off between both objectives. We evaluate CERA in two settings, $\lambda \in \{0.01, 0.05\}$, to assess the model's sensitivity to different levels of alignment strength.

\section{Experimental Setup}
\label{sec:experimentalsetup}
\subsection{Dataset Pre-Processing}
We split the 3,393 clinical trial documents into training (80\%) and test (20\%) sets using random shuffling with a fixed seed to ensure reproducibility, as shown in Table \ref{tab:dataset_split}. For each query-document pair (and 1916 queries), the document is segmented into fixed-length chunks, as shown in Equation \ref{eq:chunk}. The chunk size was determined by a heuristic based on a simple statistical criterion derived from the distribution of annotation lengths, where $L$ denotes the distribution of annotation lengths (measured in words), and $\sigma(L)$ denotes its standard deviation. 

\begin{equation}
\scriptsize
\text{chunk\_size} \approx \operatorname{median}(L) + \sigma(L)
\label{eq:chunk}
\end{equation}

In addition, gold evidence annotations (rationales) are segmented into token-level spans. Chunks overlapping the gold evidence are labeled as positive instances and retain the evidence content, while the remaining chunks are treated as candidate negatives. We select the top-5 most subjective chunks as hard negatives, using scores computed with the TextBlob Python library\footnote{\url{https://textblob.readthedocs.io/en/dev/}}. The total training triplets are 13,004. The train--test split breakdown is presented in Table~\ref{tab:dataset_split}. 

\begin{table}[!htb]
\centering
\scalefont{0.75}
\begin{tabular}{p{0.7cm} p{1.0cm} p{1.0cm} p{1.0cm} p{1.0cm} p{0.3cm}}
\hline
\textbf{Type} & \textbf{Text Files} & \textbf{Positive Triplets} & \textbf{Negative Triplets} & \textbf{Total Triplets} & \textbf{\%} \\
\hline
Train & 2,714 & ~1.730 & ~8,650 & 10,383 & 80 \\
Test  & 679   & 437 & 2,184 &  2,621 & 20 \\
\hline
Total & 3,393 & 2,167 & 10,834 & 13,004 & 100 \\
\hline
\end{tabular}
\caption{Distribution of files and triplets training instances in train and test splits.}
\label{tab:dataset_split}
\end{table}

\subsection{Model Architecture and Settings}
All experiments use the pretrained Contriever dense retriever initialized from the $\texttt{facebook/contriever}$\footnote{\url{https://huggingface.co/facebook/contriever}} checkpoint. The model is based on a Transformer encoder, where query and chunk embeddings are obtained from the final hidden-state representation of the \texttt{[CLS]} token. The resulting embeddings are L2-normalized prior to cosine similarity computation.

The model was fine-tuned in a triplet-based retrieval setting, with a query, a positive chunk, and multiple negative chunks. Training was performed with a maximum sequence length of 128 tokens, a batch size of 8, and 10 training epochs. Optimization was carried out using AdamW with a learning rate of $1 \times 10^{-6}$. The retrieval objective was optimized using a cosine-based triplet loss with a margin of $0.2$. In addition, an auxiliary rationale-guided alignment objective was incorporated through a KL divergence loss applied over the final-layer \texttt{[CLS]} attention distribution. The rationale alignment weight was set to either $\lambda = 0.05$ or $0.01$, while the numerical stability constant used during normalization was defined as $\epsilon = 1 \times 10^{-8}$.

Optimization used a linear learning-rate scheduler with 10\% warmup steps. For reproducibility, all experiments fixed the random seed to 51 across Python, NumPy, and PyTorch, with deterministic CuDNN execution enabled.

\subsection{Evaluation}
\label{sec:evaluation}


\textbf{Retrieval}: We evaluate retrieval performance using Recall@K (K = 1,3,5,10,50), Precision@K, Normalized Discounted Cumulative Gain (NDCG@K), Mean Average Precision (MAP@K), and Mean Reciprocal Rank (MRR@K), as shown in Equations \ref{equation_retrieval_metrics}, \ref{equation_ndcg_map_ndcg}, and \ref{equation_mrr}. In this paper, we present a \textit{Local Evaluation} approach that adopts a local configuration in which the test collection is organized at the PMCID document level. Documents are first divided into chunks, where those intersecting the annotated evidence spans are labeled as positives, while the remaining chunks are treated as negatives. All chunks associated with the same PMCID were grouped, with duplicates removed and their original order preserved. For each query, both positive and negative chunks were restricted to the local pool of the corresponding PMCID document. We use cosine similarity between normalized CLS embeddings, and the resulting rankings were used to calculate the retrieval metrics. This setup evaluates the model’s ability to identify relevant evidence chunks within the same document, emphasizing fine-grained retrieval performance.

\begin{equation}
\scriptsize
\begin{aligned}
\text{Recall@K} &= \frac{|Rel \cap TopK|}{|Rel|}
&
\text{Precision@K} &= \frac{|Rel \cap TopK|}{K}
\end{aligned}
\label{equation_retrieval_metrics}
\end{equation}

\begin{equation}
\scriptsize
\begin{aligned}
\text{NDCG@K} &= \frac{DCG@K}{IDCG@K}
&
DCG@K &= \sum_{i=1}^{K} \frac{2^{rel_i}-1}{\log_2(i+1)}
\\
\text{MAP} &= \frac{1}{|Q|} \sum_{q=1}^{|Q|} AP(q)
&
AP(q) &= \frac{1}{R_q}
\sum_{k=1}^{n} P@k \cdot rel(k)
\end{aligned}
\label{equation_ndcg_map_ndcg}
\end{equation}

\begin{equation}
\scriptsize
MRR = \frac{1}{|Q|} \sum_{i=1}^{|Q|} \frac{1}{rank_i}
\label{equation_mrr}
\end{equation}





\noindent\textbf{Explainability}: We evaluate CERA’s explanations using established metrics from the interpretability literature \cite{deyoung-etal-2020-eraser}: \textit{plausibility} via token-level IOU F1 and Token-F1, which measures human-alignment of generated explanations (consider Equations \ref{equation_iou-f1} and \ref{equation_token-f1}); as well as \textit{faithfulness} through \textit{comprehensiveness} and \textit{sufficiency} in Equations \ref{eq:comprehensiveness} and \ref{eq:sufficiency}, which assesses whether the explanation reflects the model's decision process.

\begin{equation}
\scriptsize
\text{IOU-F1} = \frac{1}{N}\sum_{i=1}^N \mathbb{I}\left(\frac{|M_i \cap H_i|}{|M_i \cup H_i|} > 0.5\right)
\label{equation_iou-f1}
\end{equation}

\begin{equation}
\scriptsize
\text{Token-F1} = \frac{1}{N} \sum_{i=1}^N 
\frac{2 |M_i \cap H_i|}{|M_i| + |H_i|}
\label{equation_token-f1}
\end{equation}

\begin{equation}
\scriptsize
\text{Comprehensiveness} =
\frac{1}{N}
\sum_{i=1}^{N}
\left(
s_j(x_i) - s_j(x_i \setminus r_i)
\right)
\label{eq:comprehensiveness}
\end{equation}

\begin{equation}
\scriptsize
\text{Sufficiency} =
\frac{1}{N}
\sum_{i=1}^{N}
\left(
s_j(x_i) - s_j(r_i)
\right)
\label{eq:sufficiency}
\end{equation}

\noindent\textbf{Factuality}: To further assess the relevance and quality of retrieved spans (including false positives that could contain factually useful information for a given query), we employ three state-of-the-art LLMs (\texttt{\small gpt-5.4-2026-03-05}, \texttt{\small mistral-large-2512}, \texttt{\small qwen3-max-2026-01-23}). The LLMs are tasked with judging the factual relevance of the retrieved spans for a given query. The full prompt and further details are found in Appendix \ref{sec:appendix_prompt}. We report average judge scores for individual retrieval ranks (R = 1,2,3) as well as jury scores from multiple LLMs, as shown in Equations \ref{equation_judge} and \ref{equation_jury}:

\begin{equation}
\scriptsize
Judge_{i}\:Score= \frac{1}{N}\sum_{j=1}^{N}s_{ij}
\label{equation_judge}
\end{equation}

\begin{equation}
\scriptsize
Jury\:Score = \frac{1}{3} \sum_{i=1}^{3}(\frac{1}{N}\sum_{j=1}^{N}s_{ij})
\label{equation_jury}
\end{equation}

\noindent where $s_{ij}$ is the score given by LLM $i$ to span $j$ ranked at $R$. The LLMs are tasked to assign scores between 0, 1, 2, and 3, where 0 means that the retrieved span is factually incorrect (even if relevant for the query) and 3 that it is factually consistent and highly useful to answer the query. In this fashion, we leverage the LLM's own factuality assessment of the retrieved spans as well as the LLM's judgment of their fit for the given query.

\section{Results and Discussion}
\label{sec:evaluation_results}

\paragraph{Hard Negative Selection by Subjectivity} Table~\ref{tab:retrieval_results} shows that CERA consistently outperforms both the base Contriever model and the hard-negative baseline across all metrics, with the largest gains concentrated in early-ranking performance. Particularly, improvements in Recall@1 (+0.0768), Precision@1 (+0.0934), MRR (+0.1009), Recall@5 (+0.1307), and NDCG@5 (+0.1134) indicate vastly improved prioritization of relevant evidence, which is especially important in RAG settings. While gains remain stable at higher cutoffs, the greatest improvements occur in early- and mid-ranking positions, suggesting that CERA enhances evidence prioritization rather than broad retrieval coverage alone. Appendix~\ref{sec:ape_corpus_analysis} presents a corpus analysis showing lexical differences between evidential and non-evidential content, supporting subjectivity-based hard negative selection to improve embedding discrimination and retrieval quality.

\noindent\textbf{Interpretable Attention Alignment and Evidential Inductive Bias} 
The results in Tables~\ref{tab:cera_weighted_comparison} and~\ref{tab:lambda_results} show that CERA substantially improves over Contriever, while the alignment objective introduces only minor retrieval variations (typically $\sim$0.01 absolute across metrics). This suggests that CERA$_{\text{alignment}}$ preserves the model’s retrieval capacity while slightly smoothing ranking distributions. Overall, CERA$_{\text{alignment}}$ maintains strong retrieval performance while improving evidence quality for trustworthy RAG systems. Tables~\ref{tab:cera_weighted_comparison} and~\ref{tab:lambda_results} show that CERA outperforms Contriever, while the alignment objective introduces only minor retrieval variations ($\sim$0.01 $\Delta$). This suggests that CERA$_{\text{alignment}}$ preserves retrieval capacity while slightly smoothing ranking distributions. Furthermore, the alignment objective consistently improves interpretability and faithfulness metrics, particularly sufficiency, indicating the model relies on more compact and adequate supporting evidence. 

\begin{table*}[!htb]
\centering
\begin{minipage}[t]{0.49\textwidth}
\centering
\scalefont{0.65}
\begin{tabular}{p{1.25cm}p{1.2cm}p{1.3cm}p{0.7cm}p{0.55cm}}
\hline
\textbf{Metrics} & 
$\mathbf{Contriever}$ & 
$\mathbf{HardNeg_{\text{base}}}$ & 
$\mathbf{CERA}$ &
$\Delta$ \\
\hline

\multicolumn{5}{l}{\textbf{Recall@K}} \\[2pt]
Recall@1  & 0.0214 & 0.1248 & 0.2016 & +0.0768 \\
Recall@3  & 0.0640 & 0.3096 & 0.4169 & +0.1073 \\
Recall@5  & 0.1169 & 0.4301 & 0.5608 & +0.1307 \\
Recall@10 & 0.2035 & 0.6356 & 0.7442 & +0.1086 \\
Recall@20 & 0.3392 & 0.8360 & 0.8947 & +0.0587 \\
Recall@50 & 0.6027 & 0.9627 & 0.9802 & +0.0175 \\
\hline
\multicolumn{5}{l}{\textbf{Precision@K}} \\[2pt]
Precision@1  & 0.0302 & 0.1659 & 0.2593 & +0.0934 \\
Precision@3  & 0.0299 & 0.1355 & 0.1863 & +0.0508 \\
Precision@5  & 0.0315 & 0.1142 & 0.1504 & +0.0362 \\
Precision@10 & 0.0277 & 0.0853 & 0.1010 & +0.0157 \\
Precision@20 & 0.0231 & 0.0564 & 0.0607 & +0.0043 \\
Precision@50 & 0.0164 & 0.0261 & 0.0266 & +0.0004\\
\hline
\multicolumn{5}{l}{\textbf{NDCG@K}} \\[2pt]
NDCG@3  & 0.0493 & 0.2461 & 0.3496 & +0.1035 \\
NDCG@5  & 0.0722 & 0.2995 & 0.4129 & +0.1134 \\
NDCG@10 & 0.1027 & 0.3717 & 0.4776 & +0.1059 \\
NDCG@20 & 0.1395 & 0.4267 & 0.5191 & +0.0924 \\
NDCG@50 & 0.1956 & 0.4547 & 0.5379 & +0.0831 \\
\hline
\multicolumn{5}{l}{\textbf{MAP@K}} \\[2pt]
MAP@1  & 0.0302 & 0.1659 & 0.2593 & +0.0934 \\
MAP@3  & 0.0387 & 0.2070 & 0.3057 & +0.0986 \\
MAP@5  & 0.0509 & 0.2381 & 0.3437 & +0.1055 \\
MAP@10 & 0.0632 & 0.2712 & 0.3745 & +0.1033 \\
MAP@20 & 0.0735 & 0.2888 & 0.3883 & +0.0994 \\
MAP@50 & 0.0831 & 0.2946 & 0.3921 & +0.0974 \\
\hline
\textbf{MRR} & 0.1063 & 0.3288 & 0.4296 & +0.10 \\
\hline
\end{tabular}
\caption{Local evaluation on the Evidence Inference Dataset 2.0 (1,916 queries). $\Delta$ denotes the performance gain of CERA over the hard-negative baseline.}
\label{tab:retrieval_results}
\end{minipage}
\hfill
\begin{minipage}[t]{0.49\textwidth}
\centering
\scalefont{0.63}
\begin{tabular}{p{1.1cm}p{1.2cm}p{0.6cm}p{1.35cm}p{1.35cm}}
\hline
\textbf{Metrics} & 
$\mathbf{Contriever}$ & 
$\mathbf{CERA}$ & 
$\mathbf{CERA_{\text{alignment}}^{\lambda=0.01}}$ &
$\mathbf{CERA_{\text{alignment}}^{\lambda=0.05}}$ \\
\hline

\multicolumn{5}{l}{\textbf{Recall@K}} \\[2pt]
Recall@1  & 0.0214 & 0.2016 & \textbf{0.1898} & 0.1747 \\
Recall@3  & 0.0640 & 0.4169 & \textbf{0.4026} & 0.3703 \\
Recall@5  & 0.1169 & 0.5608 & \textbf{0.5391} & 0.5085 \\
Recall@10 & 0.2035 & 0.7442 & \textbf{0.7345} & 0.7110 \\
Recall@20 & 0.3392 & 0.8947 & \textbf{0.8880} & 0.8816 \\
Recall@50 & 0.6027 & 0.9802 & \textbf{0.9779} & 0.9789 \\

\hline
\multicolumn{5}{l}{\textbf{Precision@K}} \\[2pt]
Precision@1  & 0.0302 & 0.2593 & \textbf{0.2473} & 0.2291 \\
Precision@3  & 0.0299 & 0.1863 & \textbf{0.1795} & 0.1651 \\
Precision@5  & 0.0315 & 0.1504 & \textbf{0.1449} & 0.1369 \\
Precision@10 & 0.0277 & 0.1010 & \textbf{0.0995} & 0.0966 \\
Precision@20 & 0.0231 & 0.0607 & \textbf{0.0603} & 0.0599 \\
Precision@50 & 0.0164 & 0.0266 & \textbf{0.0265} & 0.0266 \\
\hline

\multicolumn{5}{l}{\textbf{NDCG@K}} \\[2pt]
NDCG@1  & 0.0302 & 0.2593 & \textbf{0.2473} & 0.2291 \\
NDCG@3  & 0.0493 & 0.3496 & \textbf{0.3357} & 0.3081 \\
NDCG@5  & 0.0722 & 0.4129 & \textbf{0.3962} & 0.3693 \\
NDCG@10 & 0.1027 & 0.4776 & \textbf{0.4647} & 0.4402 \\
NDCG@20 & 0.1395 & 0.5191 & \textbf{0.5073} & 0.4875 \\
NDCG@50 & 0.1956 & 0.5379 & \textbf{0.5270} & 0.5087 \\

\hline
\multicolumn{5}{l}{\textbf{MAP@K}} \\[2pt]
MAP@1  & 0.0302 & 0.2593 & \textbf{0.2473} & 0.2291 \\
MAP@3  & 0.0387 & 0.3057 & \textbf{0.2926} & 0.2668 \\
MAP@5  & 0.0509 & 0.3437 & \textbf{0.3287} & 0.3035 \\
MAP@10 & 0.0632 & 0.3745 & \textbf{0.3611} & 0.3367 \\
MAP@20 & 0.0735 & 0.3883 & \textbf{0.3752} & 0.3522 \\
MAP@50 & 0.0831 & 0.3921 & \textbf{0.3792} & 0.3565 \\

\hline
\textbf{MRR} & 0.1063 & 0.4296 & 0.4166 & 0.3934 \\
\hline
\end{tabular} 
\caption{Retrieval comparison between Contriever, CERA, and CERA$_{\text{alignment}}$ variants in the ablation study of different $\lambda$ values.}
\label{tab:cera_weighted_comparison} 
\end{minipage}
\end{table*}

\begin{table*}[!htb]
\centering
\footnotesize
\scalefont{0.78}
\setlength{\tabcolsep}{4pt}
\renewcommand{\arraystretch}{1.1}
\begin{tabular}{lcccccc}
\hline
& \multicolumn{4}{c}{\textbf{Plausibility}} & \multicolumn{2}{c}{\textbf{Faithfulness}} \\
\cline{2-5} \cline{6-7}
Model & IOU-F1 $\uparrow$ & Token P $\uparrow$ & Token R $\uparrow$ & Token F1 $\uparrow$ & Comp. $\uparrow$ & Suff. $\downarrow$ \\
\hline
Contriever-base & 0.1393 & 0.7613 & 0.4037 & 0.5204 & \textbf{0.1729} & 0.4273 \\
CERA$_{\text{}}$ & 0.1514 & 0.7637 & 0.4078 & 0.5242 & 0.1056 & 0.2073 \\
CERA$_{\text{alignment}}$  $\lambda$ 0.01  & 0.1570 & \textbf{0.7637} & 0.4131 & \textbf{0.5286} & 0.1173 & 0.1250 \\
CERA$_{\text{alignment}}$ $\lambda$ 0.05  & \textbf{0.1701} & 0.7568 & \textbf{0.4163} & 0.5283 & 0.1113 & \textbf{0.0939} \\
\hline
\end{tabular}
\caption{Evaluation results across interpretability (Plausibility and Faithfulness) metrics.}
\label{tab:lambda_results}
\end{table*}


\begin{table*}[!htb]
\centering
\scalefont{0.59}
\begin{tabular}{
p{2.0cm}|
p{0.5cm}|
p{0.9cm}p{1.3cm}|
p{0.9cm}p{1.3cm}|
p{0.9cm}p{1.3cm}|
p{0.9cm}p{1.3cm}
}
\hline
\multirow{2}{*}{\textbf{Model}} &
\multirow{2}{*}{\textbf{Epoch}} &
\multicolumn{2}{c|}{\textbf{Recall@10}} &
\multicolumn{2}{c|}{\textbf{NDCG@10}} &
\multicolumn{2}{c|}{\textbf{MAP@10}} &
\multicolumn{2}{c}{\textbf{MRR}} \\

& 
& CERA$_{\text{alignment}}$ & HardNeg$_{\text{base}}$
& CERA$_{\text{alignment}}$ & HardNeg$_{\text{base}}$
& CERA$_{\text{alignment}}$ & HardNeg$_{\text{base}}$
& CERA$_{\text{alignment}}$ & HardNeg$_{\text{base}}$ \\
\hline

batch8\_2e-6
& 1
& 0.70129 & 0.63948
& 0.42486 & 0.37716
& 0.31915 & 0.27804
& 0.37729 & 0.33334 \\

batch8\_1e-6
& 2
& 0.70198 & 0.64205
& 0.42427 & 0.37845
& 0.31826 & 0.27903
& 0.37572 & 0.33370 \\

batch4\_1e-6
& 2 / 1
& 0.68515 & 0.62709
& 0.41619 & 0.37125
& 0.31333 & 0.27406
& 0.37118 & 0.33130 \\

batch4\_1e-6\_sched
& 3
& 0.68646 & 0.64144
& 0.42127 & 0.37543
& 0.31955 & 0.27428
& 0.37725 & 0.33203 \\

batch8\_1e-6\_sched
& 4 / 3
& 0.70424 & 0.63683
& 0.42823 & 0.37615
& 0.32350 & 0.27732
& 0.37905 & 0.33301 \\

batch16\_1e-6\_sched
& 8 / 4
& 0.69046 & 0.64135
& 0.42689 & 0.37612
& 0.32535 & 0.27622
& 0.38419 & 0.33084 \\

batch32\_1e-6\_sched
& 10 / 8
& 0.70390 & 0.64266
& 0.42455 & 0.37812
& 0.31822 & 0.27805
& 0.37491 & 0.33286 \\

\hline
\end{tabular}
\caption{Comparison between CERA$_{\text{alignment}}$ ($\lambda=0.05$) and HardNeg$_{\text{base}}$ across different training configurations.}
\label{tab:cera_vs_hardneg_app}
\end{table*}

\paragraph{Factuality Evaluation} The results in Table~\ref{tab:cera_factuality_llm_judges} evaluate the factuality of retrieved evidence spans produced by Contriever and CERA using multiple LLM judges (GPT5.4, Qwen3-Max, and Mistral Large 3) and an aggregated jury score. Across all judges and ranks, CERA consistently outperforms the Contriever baseline, indicating improved factuality and query grounding of retrieved evidence. Gains are observed across Rank@1–3, with particularly strong improvements under Mistral Large 3, suggesting robustness across different evaluators. The aggregated jury scores confirm this trend, with CERA achieving higher factuality across all ranks. Overall, the results indicate that subjectivity-based hard negative selection improves the factual quality of retrieved evidence.

\begin{table*}[!htb]
\centering
\scalefont{0.55}
\begin{tabular}{
p{2.2cm}|
p{0.9cm}p{0.9cm}|
p{0.9cm}p{0.9cm}|
p{0.9cm}p{0.9cm}
}
\hline
\multirow{2}{*}{\textbf{Judge}} &
\multicolumn{2}{c|}{\textbf{Rank 1}} &
\multicolumn{2}{c|}{\textbf{Rank 2}} &
\multicolumn{2}{c}{\textbf{Rank 3}} \\
& \textbf{Contriever} & \textbf{CERA} &
\textbf{Contriever} & \textbf{CERA} &
\textbf{Contriever} & \textbf{CERA} \\
\hline
GPT5.4
& 1.4379 & 2.0778 & 1.4489 & 1.9671 & 1.4207 & 1.8737 \\
Qwen3-Max
& 1.1957 & 1.7724 & 1.2114 & 1.6059 & 1.1696 & 1.4703 \\
Mistral Large 3
& 0.9306 & 1.8544 & 0.8914 & 1.6628 & 0.8199 & 1.5016 \\
\hline
Jury
& 1.1881 & 1.9015 & 1.1839 & 1.7453 & 1.1367 & 1.6152 \\
\hline
\end{tabular}
\caption{Factuality evaluation comparing Contriever and CERA across LLM judges and ranks. Each rank reports side-by-side scores for both models.}
\label{tab:cera_factuality_llm_judges}
\end{table*}

\subsection{Ablation Studies}
\paragraph{Sensitivity Analysis of the Alignment Weight $\lambda$}Tables~\ref{tab:cera_alignment_variants_app} and~\ref{tab:cera_weighted_comparison} show that increasing the alignment weight from $\lambda=0.01$ to $\lambda=0.05$ reduces retrieval performance, particularly in early-ranking metrics such as Recall@1, NDCG@5, MAP@10, and MRR. While larger retrieval cutoffs remain relatively stable, the results suggest that stronger alignment supervision mainly affects fine-grained ranking quality. Overall, moderate alignment strength provides a better balance between retrieval effectiveness and rationale-guided supervision.

\noindent\textbf{Training Configuration and Hyperparameter Sensitivity Analysis.}
Table~\ref{tab:cera_vs_hardneg_app} presents an ablation study comparing different training configurations of  CERA$_{\text{alignment}}$ against a hard negative selection baseline across varying batch sizes, learning rates, optimizers\footnote{\url{https://huggingface.co/transformers/v4.2.2/main_classes/optimizer_schedules.html}}, and training epochs. CERA$_{\text{alignment}}$ consistently outperforms HardNeg$_{\text{base}}$, with the largest gains observed in Recall@10 (approximately +5--7 absolute points). Similar improvements in NDCG@10, MAP@10, and MRR indicate better ranking quality and early precision. The best overall configurations are batch8\_1e-6\_sched for Recall@10 and NDCG@10, and batch16\_1e-6\_sched for MAP@10 and MRR. Results suggest that subjectivity-based negatives provide a more informative contrastive signal than conventional hard negatives, yielding more discriminative retrieval representations. A more detailed ablation study analysis is provided in the Appendix \ref{sec:appendix_ablation}.


\section{Conclusions}

This paper introduced Contrastive Evidence Rationale Attention (CERA), the first retrieval framework that moves RAG beyond topical similarity toward evidence-aware and interpretable retrieval. By combining subjectivity-based hard negative selection with rationale-guided attention alignment, CERA injects an evidential inductive bias into contrastive retrieval training. Experiments on the Evidence Inference 2.0 dataset demonstrate consistent improvements in retrieval effectiveness across Recall, NDCG, MAP, and MRR metrics, with the best configuration improving Recall@10 from 0.63566 to 0.74426 and MRR from 0.32880 to 0.42965 over both Contriever and hard-negative selection baselines. Results further show that rationale alignment improves interpretability while preserving competitive retrieval performance. Overall, our findings suggest that retrieval systems should optimize not only semantic relevance, but also evidential support and interpretability, particularly in high-stakes domains where trustworthy and interpretable evidence selection is critical.

\section{Limitations}
Despite promising results, this work has limitations. First, experiments are restricted to the Evidence Inference 2.0 dataset, limiting generalization to other domains and retrieval settings. Second, the alignment objective depends on expert-annotated rationales, which are costly and possibly subjective. Third, attention weights may not always reflect model reasoning, despite improvements in plausibility and faithfulness. Fourth, only a limited set of alignment hyperparameters is evaluated; multi-hop reasoning and long-range dependencies are not addressed. Finally, CERA should not be used as a standalone clinical decision-making system without human oversight.

\section{Ethics Statement}
This work improves interpretable evidence retrieval for biomedical NLP using publicly available clinical trial data, with no patient interaction or personal health data involved. Although rationale alignment improves interpretability, attention-based explanations do not guarantee factual correctness or clinical reliability, and the model may inherit biases from the underlying data. Thus, CERA is a research prototype and not a replacement for medical judgment. We hope this work leads to more transparent and responsible evidence retrieval systems.

\section*{Acknowledgments}
The authors are grateful to the São Paulo Research Foundation (FAPESP) (grants \#2025/01118-2 and \#2024/04890-5) for partial financial support. The authors also thank Dr. Manuel Carrillo (São Paulo State University) and Dr. Jackson Trager (University of Southern California) for their insightful discussions and valuable support, particularly regarding the review and assessment of factuality, plausibility, and faithfulness metrics. Finally, part of this research was conducted during the first author's postdoctoral research at São Paulo State University, in collaboration with Idiap Research Institute.

\bibliography{custom}

@inproceedings{wiegreffe-pinter-2019-attention,
    title = "Attention is not not Explanation",
    author = "Wiegreffe, Sarah  and
      Pinter, Yuval",
    editor = "Inui, Kentaro  and
      Jiang, Jing  and
      Ng, Vincent  and
      Wan, Xiaojun",
    booktitle = "Proceedings of the 2019 Conference on Empirical Methods in Natural Language Processing and the 9th International Joint Conference on Natural Language Processing (EMNLP-IJCNLP)",
    month = nov,
    year = "2019",
    address = "Hong Kong, China",
    publisher = "Association for Computational Linguistics",
    url = "https://aclanthology.org/D19-1002/",
    doi = "10.18653/v1/D19-1002",
    pages = "11--20"
}

@inproceedings{nguyen-etal-2023-passage,
    title = "Passage-based {BM}25 Hard Negatives: A Simple and Effective Negative Sampling Strategy For Dense Retrieval",
    author = "Nguyen, Thanh-Do  and
      Bui, Chi Minh  and
      Vuong, Thi-Hai-Yen  and
      Phan, Xuan-Hieu",
    editor = "Huang, Chu-Ren  and
      Harada, Yasunari  and
      Kim, Jong-Bok  and
      Chen, Si  and
      Hsu, Yu-Yin  and
      Chersoni, Emmanuele  and
      A, Pranav  and
      Zeng, Winnie Huiheng  and
      Peng, Bo  and
      Li, Yuxi  and
      Li, Junlin",
    booktitle = "Proceedings of the 37th Pacific Asia Conference on Language, Information and Computation",
    month = dec,
    year = "2023",
    address = "Hong Kong, China",
    publisher = "Association for Computational Linguistics",
    url = "https://aclanthology.org/2023.paclic-1.59/",
    pages = "591--599"
}

@inproceedings{deyoung-etal-2020-eraser,
    title = "{ERASER}: {A} Benchmark to Evaluate Rationalized {NLP} Models",
    author = "DeYoung, Jay  and
      Jain, Sarthak  and
      Rajani, Nazneen Fatema  and
      Lehman, Eric  and
      Xiong, Caiming  and
      Socher, Richard  and
      Wallace, Byron C.",
    editor = "Jurafsky, Dan  and
      Chai, Joyce  and
      Schluter, Natalie  and
      Tetreault, Joel",
    booktitle = "Proceedings of the 58th Annual Meeting of the Association for Computational Linguistics",
    month = jul,
    year = "2020",
    address = "Online",
    publisher = "Association for Computational Linguistics",
    url = "https://aclanthology.org/2020.acl-main.408/",
    doi = "10.18653/v1/2020.acl-main.408",
    pages = "4443--4458"
}

@inproceedings{abolghasemi-etal-2025-evaluation,
    title = "Evaluation of Attribution Bias in Generator-Aware Retrieval-Augmented Large Language Models",
    author = "Abolghasemi, Amin  and
      Azzopardi, Leif  and
      Hashemi, Seyyed Hadi  and
      de Rijke, Maarten  and
      Verberne, Suzan",
    editor = "Che, Wanxiang  and
      Nabende, Joyce  and
      Shutova, Ekaterina  and
      Pilehvar, Mohammad Taher",
    booktitle = "Findings of the Association for Computational Linguistics: ACL 2025",
    month = jul,
    year = "2025",
    address = "Vienna, Austria",
    publisher = "Association for Computational Linguistics",
    url = "https://aclanthology.org/2025.findings-acl.1087/",
    doi = "10.18653/v1/2025.findings-acl.1087",
    pages = "21105--21124",
    ISBN = "979-8-89176-256-5"
}

@inproceedings{sun-etal-2025-fact,
    title = "Fact-Aware Multimodal Retrieval Augmentation for Accurate Medical Radiology Report Generation",
    author = "Sun, Liwen  and
      Zhao, James Jialun  and
      Han, Wenjing  and
      Xiong, Chenyan",
    editor = "Chiruzzo, Luis  and
      Ritter, Alan  and
      Wang, Lu",
    booktitle = "Proceedings of the 2025 Conference of the Nations of the Americas Chapter of the Association for Computational Linguistics: Human Language Technologies (Volume 1: Long Papers)",
    month = apr,
    year = "2025",
    address = "Albuquerque, New Mexico",
    publisher = "Association for Computational Linguistics",
    url = "https://aclanthology.org/2025.naacl-long.28/",
    doi = "10.18653/v1/2025.naacl-long.28",
    pages = "643--655",
    ISBN = "979-8-89176-189-6"
}

@misc{petroni2020contextaffectslanguagemodels,
      title={How Context Affects Language Models' Factual Predictions}, 
      author={Fabio Petroni and Patrick Lewis and Aleksandra Piktus and Tim Rocktäschel and Yuxiang Wu and Alexander H. Miller and Sebastian Riedel},
      year={2020},
      eprint={2005.04611},
      archivePrefix={arXiv},
      primaryClass={cs.CL},
      url={https://arxiv.org/abs/2005.04611}, 
}

@techreport{mitchell1980bias,
  author      = {Tom M. Mitchell},
  title       = {The Need for Biases in Learning Generalizations},
  institution = {Rutgers University, Department of Computer Science},
  number      = {CBM-TR-117},
  year        = {1980},
  url = {https://www.cs.cmu.edu/~tom/pubs/NeedForBias_1980.pdf}
}

@inproceedings{xin-etal-2025-sparse,
    title = "Sparse Latents Steer Retrieval-Augmented Generation",
    author = "Xin, Chunlei  and
      Zhou, Shuheng  and
      Zhu, Huijia  and
      Wang, Weiqiang  and
      Chen, Xuanang  and
      Guan, Xinyan  and
      Lu, Yaojie  and
      Lin, Hongyu  and
      Han, Xianpei  and
      Sun, Le",
    editor = "Che, Wanxiang  and
      Nabende, Joyce  and
      Shutova, Ekaterina  and
      Pilehvar, Mohammad Taher",
    booktitle = "Proceedings of the 63rd Annual Meeting of the Association for Computational Linguistics (Volume 1: Long Papers)",
    year = "2025",
    address = "Vienna, Austria",
    publisher = "Association for Computational Linguistics",
    url = "https://aclanthology.org/2025.acl-long.228/",
    doi = "10.18653/v1/2025.acl-long.228",
    pages = "4547--4562",
    ISBN = "979-8-89176-251-0"
}

@misc{Vashishth2019AttentionIA,
  title={Attention Interpretability Across NLP Tasks},
  author={Shikhar Vashishth and Shyam Upadhyay and Gaurav Singh Tomar and Manaal Faruqui},
  year={2019},
  eprint={1909.11218},
  archivePrefix={arXiv},
  url={https://arxiv.org/abs/1909.11218}
}

@inproceedings{serrano-smith-2019-attention,
    title = "Is Attention Interpretable?",
    author = "Serrano, Sofia  and
      Smith, Noah A.",
    editor = "Korhonen, Anna  and
      Traum, David  and
      M{\`a}rquez, Llu{\'i}s",
    booktitle = "Proceedings of the 57th Annual Meeting of the Association for Computational Linguistics",
    month = jul,
    year = "2019",
    address = "Florence, Italy",
    publisher = "Association for Computational Linguistics",
    url = "https://aclanthology.org/P19-1282/",
    doi = "10.18653/v1/P19-1282",
    pages = "2931--2951"
}

@article{Eilertsen_Bjrgfinsdttir_Vargas_Ramezani-Kebrya_2026, 
title={Aligning Attention with Human Rationales for Self-Explaining Hate Speech Detection}, volume={40}, url={https://ojs.aaai.org/index.php/AAAI/article/view/41069}, DOI={10.1609/aaai.v40i44.41069}, abstractNote={The opaque nature of deep learning models presents significant challenges for the ethical deployment of hate speech detection systems. To address this limitation, we introduce Supervised Rational Attention (SRA), a framework that explicitly aligns model attention with human rationales, improving both interpretability and fairness in hate speech classification. SRA integrates a supervised attention mechanism into transformer-based classifiers, optimizing a joint objective that combines standard classification loss with an alignment loss term that minimizes the discrepancy between attention weights and human-annotated rationales.
We evaluated SRA on hate speech benchmarks in English (HateXplain) and Portuguese (HateBRXplain) with rationale annotations. Empirically, SRA achieves 2.4× better explainability compared to current baselines, and produces token-level explanations that are more faithful and human-aligned. In terms of fairness, SRA achieves competitive fairness across all measures, with second-best performance in detecting toxic posts targeting identity groups, while maintaining comparable results on other metrics. These findings demonstrate that incorporating human rationales into attention mechanisms can enhance interpretability and faithfulness without compromising fairness.}, number={44}, journal={Proceedings of the AAAI Conference on Artificial Intelligence}, author={Eilertsen, Brage and Bjørgfinsdóttir, Røskva and Vargas, Francielle and Ramezani-Kebrya, Ali}, year={2026}, month={Mar.}, pages={37369-37378}}

@inproceedings{vargas-etal-2026-self-explaining,
    title = {Self-Explaining Hate Speech Detection with Moral Rationales},
    author = {Vargas, Francielle and Trager, Jackson and Alves, Diego and Thapa, Surendrabikram and Guida, Matteo and Atil, Berk and Dementieva, Daryna and Smart, Andrew and Agrawal, Ameeta},
    booktitle = {Findings of the Association for Computational Linguistics: ACL 2026},
    pages = {1--18},
    address = {San Diego, USA},
    publisher = {Association for Computational Linguistics},
    year = {2026},
    url = {https://arxiv.org/abs/2601.03481}
}

@inproceedings{chrysostomou-aletras-2021-improving,
    title = "Improving the Faithfulness of Attention-based Explanations with Task-specific Information for Text Classification",
    author = "Chrysostomou, George  and
      Aletras, Nikolaos",
    editor = "Zong, Chengqing  and
      Xia, Fei  and
      Li, Wenjie  and
      Navigli, Roberto",
    booktitle = "Proceedings of the 59th Annual Meeting of the Association for Computational Linguistics and the 11th International Joint Conference on Natural Language Processing (Volume 1: Long Papers)",
    month = aug,
    year = "2021",
    address = "Online",
    publisher = "Association for Computational Linguistics",
    url = "https://aclanthology.org/2021.acl-long.40/",
    doi = "10.18653/v1/2021.acl-long.40",
    pages = "477--488"
}

@article{kullback1951information,
  title={On Information and Sufficiency},
  author={Kullback, Solomon and Leibler, Richard A.},
  journal={The Annals of Mathematical Statistics},
  volume={22},
  number={1},
  pages={79--86},
  year={1951},
  publisher={Institute of Mathematical Statistics}
}

@inproceedings{kim-lee-2024-rag,
    title = "{RE}-{RAG}: Improving Open-Domain {QA} Performance and Interpretability with Relevance Estimator in Retrieval-Augmented Generation",
    author = "Kim, Kiseung  and
      Lee, Jay-Yoon",
    editor = "Al-Onaizan, Yaser  and
      Bansal, Mohit  and
      Chen, Yun-Nung",
    booktitle = "Proceedings of the 2024 Conference on Empirical Methods in Natural Language Processing",
    year = "2024",
    address = "Miami, Florida, USA",
    publisher = "Association for Computational Linguistics",
    url = "https://aclanthology.org/2024.emnlp-main.1236/",
    doi = "10.18653/v1/2024.emnlp-main.1236",
    pages = "22149--22161"
}

@inproceedings{gu-etal-2025-toward,
    title = "Toward Structured Knowledge Reasoning: Contrastive Retrieval-Augmented Generation on Experience",
    author = "Gu, Jiawei  and
      Xian, Ziting  and
      Xie, Yuanzhen  and
      Liu, Ye  and
      Liu, Enjie  and
      Zhong, Ruichao  and
      Gao, Mochi  and
      Tan, Yunzhi  and
      Hu, Bo  and
      Li, Zang",
    editor = "Che, Wanxiang  and
      Nabende, Joyce  and
      Shutova, Ekaterina  and
      Pilehvar, Mohammad Taher",
    booktitle = "Findings of the Association for Computational Linguistics: ACL 2025",
    month = jul,
    year = "2025",
    address = "Vienna, Austria",
    publisher = "Association for Computational Linguistics",
    url = "https://aclanthology.org/2025.findings-acl.1224/",
    doi = "10.18653/v1/2025.findings-acl.1224",
    pages = "23891--23910",
    ISBN = "979-8-89176-256-5"
}

@inproceedings{niu-etal-2024-ragtruth,
    title = "{RAGT}ruth: A Hallucination Corpus for Developing Trustworthy Retrieval-Augmented Language Models",
    author = "Niu, Cheng  and
      Wu, Yuanhao  and
      Zhu, Juno  and
      Xu, Siliang  and
      Shum, KaShun  and
      Zhong, Randy  and
      Song, Juntong  and
      Zhang, Tong",
    editor = "Ku, Lun-Wei  and
      Martins, Andre  and
      Srikumar, Vivek",
    booktitle = "Proceedings of the 62nd Annual Meeting of the Association for Computational Linguistics (Volume 1: Long Papers)",
    month = aug,
    year = "2024",
    address = "Bangkok, Thailand",
    publisher = "Association for Computational Linguistics",
    url = "https://aclanthology.org/2024.acl-long.585/",
    doi = "10.18653/v1/2024.acl-long.585",
    pages = "10862--10878"
}

@inproceedings{lehman-etal-2019-inferring,
    title = "Inferring Which Medical Treatments Work from Reports of Clinical Trials",
    author = "Lehman, Eric  and
      DeYoung, Jay  and
      Barzilay, Regina  and
      Wallace, Byron C.",
    editor = "Burstein, Jill  and
      Doran, Christy  and
      Solorio, Thamar",
    booktitle = "Proceedings of the 2019 Conference of the North {A}merican Chapter of the Association for Computational Linguistics: Human Language Technologies, Volume 1 (Long and Short Papers)",
    month = jun,
    year = "2019",
    address = "Minneapolis, Minnesota",
    publisher = "Association for Computational Linguistics",
    url = "https://aclanthology.org/N19-1371/",
    doi = "10.18653/v1/N19-1371",
    pages = "3705--3717"
}

@inproceedings{ranaldi-etal-2025-eliciting,
    title = "Eliciting Critical Reasoning in Retrieval-Augmented Generation via Contrastive Explanations",
    author = "Ranaldi, Leonardo  and
      Valentino, Marco  and
      Freitas, Andre",
    editor = "Chiruzzo, Luis  and
      Ritter, Alan  and
      Wang, Lu",
    booktitle = "Proceedings of the 2025 Conference of the Nations of the Americas Chapter of the Association for Computational Linguistics: Human Language Technologies (Volume 1: Long Papers)",
    month = apr,
    year = "2025",
    address = "Albuquerque, New Mexico",
    publisher = "Association for Computational Linguistics",
    url = "https://aclanthology.org/2025.naacl-long.557/",
    doi = "10.18653/v1/2025.naacl-long.557",
    pages = "11168--11183",
    ISBN = "979-8-89176-189-6"
}

@inproceedings{shi2023,
author = {Shi, Freda and Chen, Xinyun and Misra, Kanishka and Scales, Nathan and Dohan, David and Chi, Ed and Sch\"{a}rli, Nathanael and Zhou, Denny},
title = {Large language models can be easily distracted by irrelevant context},
year = {2023},
publisher = {JMLR.org},
booktitle = {Proceedings of the 40th International Conference on Machine Learning},
articleno = {1291},
numpages = {18},
location = {Honolulu, Hawaii, USA},
series = {ICML'23},
url = {https://dl.acm.org/doi/10.5555/3618408.3619699}
}

@inproceedings{10.1145/3746252.3761254,
author = {Moreira, Gabriel de Souza P. and Osmulski, Radek and Xu, Mengyao and Ak, Ronay and Schifferer, Benedikt and Oldridge, Even},
title = {Improving Text Embedding Models with Positive-aware Hard-negative Mining},
year = {2025},
isbn = {9798400720406},
publisher = {Association for Computing Machinery},
address = {New York, NY, USA},
url = {https://doi.org/10.1145/3746252.3761254},
doi = {10.1145/3746252.3761254},
booktitle = {Proceedings of the 34th ACM International Conference on Information and Knowledge Management},
pages = {2169–2178},
numpages = {10},
location = {Seoul, Republic of Korea},
series = {CIKM '25}
}

@misc{saxena2025rankingfreerag,
      title={Ranking Free RAG: Replacing Re-ranking with Selection in RAG for Sensitive Domains}, 
      author={Yash Saxena and Ankur Padia and Mandar S Chaudhary and Kalpa Gunaratna and Srinivasan Parthasarathy and Manas Gaur},
      year={2025},
      eprint={2505.16014},
      archivePrefix={arXiv},
      primaryClass={cs.CL},
      note={Version 3},
      url={https://arxiv.org/abs/2505.16014}, 
}

@inproceedings{lundberg2017,
author = {Lundberg, Scott M. and Lee, Su-In},
title = {A unified approach to interpreting model predictions},
year = {2017},
isbn = {9781510860964},
publisher = {Curran Associates Inc.},
address = {Red Hook, NY, USA},
booktitle = {Proceedings of the 31st International Conference on Neural Information Processing Systems},
pages = {4768–4777},
numpages = {10},
location = {Long Beach, California, USA},
series = {NIPS'17},
url = {https://dl.acm.org/doi/10.5555/3295222.3295230}
}

@inproceedings{10.1145/2939672.2939778,
author = {Ribeiro, Marco Tulio and Singh, Sameer and Guestrin, Carlos},
title = {"Why Should I Trust You?": Explaining the Predictions of Any Classifier},
year = {2016},
isbn = {9781450342322},
publisher = {Association for Computing Machinery},
address = {New York, NY, USA},
url = {https://doi.org/10.1145/2939672.2939778},
doi = {10.1145/2939672.2939778},
booktitle = {Proceedings of the 22nd ACM SIGKDD International Conference on Knowledge Discovery and Data Mining},
pages = {1135–1144},
numpages = {10},
location = {San Francisco, California, USA},
series = {KDD '16}
}

@inproceedings{deng-etal-2024-learning,
    title = "Learning Interpretable Legal Case Retrieval via Knowledge-Guided Case Reformulation",
    author = "Deng, Chenlong  and
      Mao, Kelong  and
      Dou, Zhicheng",
    editor = "Al-Onaizan, Yaser  and
      Bansal, Mohit  and
      Chen, Yun-Nung",
    booktitle = "Proceedings of the 2024 Conference on Empirical Methods in Natural Language Processing",
    month = nov,
    year = "2024",
    address = "Miami, Florida, USA",
    publisher = "Association for Computational Linguistics",
    url = "https://aclanthology.org/2024.emnlp-main.73/",
    doi = "10.18653/v1/2024.emnlp-main.73",
    pages = "1253--1265"
}

@inproceedings{li-etal-2025-iris,
    title = "{IRIS}: Interpretable Retrieval-Augmented Classification for Long Interspersed Document Sequences",
    author = "Li, Fengnan  and
      Hill, Elliot D.  and
      Shu, Jiang  and
      Gao, Jiaxin  and
      Engelhard, Matthew M.",
    editor = "Che, Wanxiang  and
      Nabende, Joyce  and
      Shutova, Ekaterina  and
      Pilehvar, Mohammad Taher",
    booktitle = "Proceedings of the 63rd Annual Meeting of the Association for Computational Linguistics (Volume 1: Long Papers)",
    month = jul,
    year = "2025",
    address = "Vienna, Austria",
    publisher = "Association for Computational Linguistics",
    url = "https://aclanthology.org/2025.acl-long.1461/",
    doi = "10.18653/v1/2025.acl-long.1461",
    pages = "30263--30283",
    ISBN = "979-8-89176-251-0"
}

@inproceedings{jacovi-etal-2021-contrastive,
    title = "Contrastive Explanations for Model Interpretability",
    author = "Jacovi, Alon  and
      Swayamdipta, Swabha  and
      Ravfogel, Shauli  and
      Elazar, Yanai  and
      Choi, Yejin  and
      Goldberg, Yoav",
    editor = "Moens, Marie-Francine  and
      Huang, Xuanjing  and
      Specia, Lucia  and
      Yih, Scott Wen-tau",
    booktitle = "Proceedings of the 2021 Conference on Empirical Methods in Natural Language Processing",
    month = nov,
    year = "2021",
    address = "Online and Punta Cana, Dominican Republic",
    publisher = "Association for Computational Linguistics",
    url = "https://aclanthology.org/2021.emnlp-main.120/",
    doi = "10.18653/v1/2021.emnlp-main.120",
    pages = "1597--1611"
}

@inproceedings{liu-etal-2023-evaluating,
    title = "Evaluating Verifiability in Generative Search Engines",
    author = "Liu, Nelson  and
      Zhang, Tianyi  and
      Liang, Percy",
    editor = "Bouamor, Houda  and
      Pino, Juan  and
      Bali, Kalika",
    booktitle = "Findings of the Association for Computational Linguistics: EMNLP 2023",
    month = dec,
    year = "2023",
    address = "Singapore",
    publisher = "Association for Computational Linguistics",
    url = "https://aclanthology.org/2023.findings-emnlp.467/",
    doi = "10.18653/v1/2023.findings-emnlp.467",
    pages = "7001--7025"
}

@article{ram-etal-2023-context,
    title = "In-Context Retrieval-Augmented Language Models",
    author = "Ram, Ori  and
      Levine, Yoav  and
      Dalmedigos, Itay  and
      Muhlgay, Dor  and
      Shashua, Amnon  and
      Leyton-Brown, Kevin  and
      Shoham, Yoav",
    journal = "Transactions of the Association for Computational Linguistics",
    volume = "11",
    year = "2023",
    address = "Cambridge, MA",
    publisher = "MIT Press",
    doi = "10.1162/tacl_a_00605",
    url = "https://aclanthology.org/2023.tacl-1.75/",
    pages = "1316--1331"
}

@inproceedings{lewis2020retrieval,
author = {Lewis, Patrick and Perez, Ethan and Piktus, Aleksandra and Petroni, Fabio and Karpukhin, Vladimir and Goyal, Naman and K\"{u}ttler, Heinrich and Lewis, Mike and Yih, Wen-tau and Rockt\"{a}schel, Tim and Riedel, Sebastian and Kiela, Douwe},
title = {Retrieval-augmented generation for knowledge-intensive NLP tasks},
year = {2020},
isbn = {9781713829546},
publisher = {Curran Associates Inc.},
address = {Red Hook, NY, USA},
articleno = {793},
numpages = {16},
location = {Vancouver, BC, Canada},
series = {NIPS '20},
url={https://dl.acm.org/doi/abs/10.5555/3495724.3496517}
}

@inproceedings{delbrouck-etal-2022-improving,
    title = "Improving the Factual Correctness of Radiology Report Generation with Semantic Rewards",
    author = "Delbrouck, Jean-Benoit  and
      Chambon, Pierre  and
      Bluethgen, Christian  and
      Tsai, Emily  and
      Almusa, Omar  and
      Langlotz, Curtis",
    editor = "Goldberg, Yoav  and
      Kozareva, Zornitsa  and
      Zhang, Yue",
    booktitle = "Findings of the Association for Computational Linguistics: EMNLP 2022",
    month = dec,
    year = "2022",
    address = "Abu Dhabi, United Arab Emirates",
    publisher = "Association for Computational Linguistics",
    url = "https://aclanthology.org/2022.findings-emnlp.319/",
    doi = "10.18653/v1/2022.findings-emnlp.319",
    pages = "4348--4360"
}

@article{Legislation2016,
author = {General Data Protection Regulation},
year = {2016},
title = {Regulation(eu) 2016/679 of the european parliament and of the council of 27 april 2016 on the protection of natural persons with regard to the processing of personal
data and on the free movement of such data, and repealing directive 95/46},
journal = {Official Journal of the European Union (OJ)},
volume = {59},
number = {1-88},
url={https://eur-lex.europa.eu/eli/reg/2016/679/oj}
}

@misc{izacard2022unsuperviseddenseinformationretrieval,
      title={Unsupervised Dense Information Retrieval with Contrastive Learning}, 
      author={Gautier Izacard and Mathilde Caron and Lucas Hosseini and Sebastian Riedel and Piotr Bojanowski and Armand Joulin and Edouard Grave},
      year={2022},
      eprint={2112.09118},
      archivePrefix={arXiv},
      primaryClass={cs.IR},
      url={https://arxiv.org/abs/2112.09118}, 
}

@inproceedings{chatzikyriakidis-natsina-2025-poetry,
    title = "Poetry in {RAG}s: {M}odern {G}reek interwar poetry generation using {RAG} and contrastive training",
    author = "Chatzikyriakidis, Stergios  and
      Natsina, Anastasia",
    editor = {H{\"a}m{\"a}l{\"a}inen, Mika  and
      {\"O}hman, Emily  and
      Bizzoni, Yuri  and
      Miyagawa, So  and
      Alnajjar, Khalid},
    booktitle = "Proceedings of the 5th International Conference on Natural Language Processing for Digital Humanities",
    month = may,
    year = "2025",
    address = "Albuquerque, USA",
    publisher = "Association for Computational Linguistics",
    url = "https://aclanthology.org/2025.nlp4dh-1.22/",
    doi = "10.18653/v1/2025.nlp4dh-1.22",
    pages = "257--264",
    ISBN = "979-8-89176-234-3"
}

@inproceedings{sriram-etal-2024-contrastive,
    title = "Contrastive Learning to Improve Retrieval for Real-World Fact Checking",
    author = "Sriram, Aniruddh  and
      Xu, Fangyuan  and
      Choi, Eunsol  and
      Durrett, Greg",
    editor = "Schlichtkrull, Michael  and
      Chen, Yulong  and
      Whitehouse, Chenxi  and
      Deng, Zhenyun  and
      Akhtar, Mubashara  and
      Aly, Rami  and
      Guo, Zhijiang  and
      Christodoulopoulos, Christos  and
      Cocarascu, Oana  and
      Mittal, Arpit  and
      Thorne, James  and
      Vlachos, Andreas",
    booktitle = "Proceedings of the Seventh Fact Extraction and VERification Workshop (FEVER)",
    month = nov,
    year = "2024",
    address = "Miami, Florida, USA",
    publisher = "Association for Computational Linguistics",
    url = "https://aclanthology.org/2024.fever-1.28/",
    doi = "10.18653/v1/2024.fever-1.28",
    pages = "264--279"
}

@inproceedings{ghate-etal-2025-biases,
    title = "Biases Propagate in Encoder-based Vision-Language Models: A Systematic Analysis From Intrinsic Measures to Zero-shot Retrieval Outcomes",
    author = "Ghate, Kshitish  and
      Charlesworth, Tessa  and
      Diab, Mona T.  and
      Caliskan, Aylin",
    editor = "Che, Wanxiang  and
      Nabende, Joyce  and
      Shutova, Ekaterina  and
      Pilehvar, Mohammad Taher",
    booktitle = "Findings of the Association for Computational Linguistics: ACL 2025",
    month = jul,
    year = "2025",
    address = "Vienna, Austria",
    publisher = "Association for Computational Linguistics",
    url = "https://aclanthology.org/2025.findings-acl.955/",
    doi = "10.18653/v1/2025.findings-acl.955",
    pages = "18562--18580",
    ISBN = "979-8-89176-256-5"
}

@inproceedings{liang-etal-2020-alice,
    title = "{ALICE}: Active Learning with Contrastive Natural Language Explanations",
    author = "Liang, Weixin  and
      Zou, James  and
      Yu, Zhou",
    editor = "Webber, Bonnie  and
      Cohn, Trevor  and
      He, Yulan  and
      Liu, Yang",
    booktitle = "Proceedings of the 2020 Conference on Empirical Methods in Natural Language Processing (EMNLP)",
    month = nov,
    year = "2020",
    address = "Online",
    publisher = "Association for Computational Linguistics",
    url = "https://aclanthology.org/2020.emnlp-main.355/",
    doi = "10.18653/v1/2020.emnlp-main.355",
    pages = "4380--4391"
}

@inproceedings{cheng-amiri-2025-equalizeir,
    title = "{E}qualize{IR}: Mitigating Linguistic Biases in Retrieval Models",
    author = "Cheng, Jiali  and
      Amiri, Hadi",
    editor = "Chiruzzo, Luis  and
      Ritter, Alan  and
      Wang, Lu",
    booktitle = "Proceedings of the 2025 Conference of the Nations of the Americas Chapter of the Association for Computational Linguistics: Human Language Technologies (Volume 2: Short Papers)",
    month = apr,
    year = "2025",
    address = "Albuquerque, New Mexico",
    publisher = "Association for Computational Linguistics",
    url = "https://aclanthology.org/2025.naacl-short.75/",
    doi = "10.18653/v1/2025.naacl-short.75",
    pages = "889--898",
    ISBN = "979-8-89176-190-2"
}

@inproceedings{wu-etal-2025-medical,
    title = "Medical Graph {RAG}: Evidence-based Medical Large Language Model via Graph Retrieval-Augmented Generation",
    author = "Wu, Junde  and
      Zhu, Jiayuan  and
      Qi, Yunli  and
      Chen, Jingkun  and
      Xu, Min  and
      Menolascina, Filippo  and
      Jin, Yueming  and
      Grau, Vicente",
    editor = "Che, Wanxiang  and
      Nabende, Joyce  and
      Shutova, Ekaterina  and
      Pilehvar, Mohammad Taher",
    booktitle = "Proceedings of the 63rd Annual Meeting of the Association for Computational Linguistics (Volume 1: Long Papers)",
    month = jul,
    year = "2025",
    address = "Vienna, Austria",
    publisher = "Association for Computational Linguistics",
    url = "https://aclanthology.org/2025.acl-long.1381/",
    doi = "10.18653/v1/2025.acl-long.1381",
    pages = "28443--28467",
    ISBN = "979-8-89176-251-0"
}

@inproceedings{kim-etal-2024-discovering,
    title = "Discovering Biases in Information Retrieval Models Using Relevance Thesaurus as Global Explanation",
    author = "Kim, Youngwoo  and
      Rahimi, Razieh  and
      Allan, James",
    editor = "Al-Onaizan, Yaser  and
      Bansal, Mohit  and
      Chen, Yun-Nung",
    booktitle = "Proceedings of the 2024 Conference on Empirical Methods in Natural Language Processing",
    month = nov,
    year = "2024",
    address = "Miami, Florida, USA",
    publisher = "Association for Computational Linguistics",
    url = "https://aclanthology.org/2024.emnlp-main.1089/",
    doi = "10.18653/v1/2024.emnlp-main.1089",
    pages = "19530--19547"
}

@inproceedings{chen-etal-2025-llms,
    title = "{LLM}s are Biased Evaluators But Not Biased for Fact-Centric Retrieval Augmented Generation",
    author = "Chen, Yen-Shan  and
      Jin, Jing  and
      Kuo, Peng-Ting  and
      Huang, Chao-Wei  and
      Chen, Yun-Nung",
    editor = "Che, Wanxiang  and
      Nabende, Joyce  and
      Shutova, Ekaterina  and
      Pilehvar, Mohammad Taher",
    booktitle = "Findings of the Association for Computational Linguistics: ACL 2025",
    year = "2025",
    address = "Vienna, Austria",
    publisher = "Association for Computational Linguistics",
    url = "https://aclanthology.org/2025.findings-acl.1369/",
    doi = "10.18653/v1/2025.findings-acl.1369",
    pages = "26669--26684",
    ISBN = "979-8-89176-256-5"
}

@inproceedings{hu-etal-2025-removal,
    title = "Removal of Hallucination on Hallucination: Debate-Augmented {RAG}",
    author = "Hu, Wentao  and
      Zhang, Wengyu  and
      Jiang, Yiyang  and
      Zhang, Chen Jason  and
      Wei, Xiaoyong  and
      Qing, Li",
    editor = "Che, Wanxiang  and
      Nabende, Joyce  and
      Shutova, Ekaterina  and
      Pilehvar, Mohammad Taher",
    booktitle = "Proceedings of the 63rd Annual Meeting of the Association for Computational Linguistics (Volume 1: Long Papers)",
    month = jul,
    year = "2025",
    address = "Vienna, Austria",
    publisher = "Association for Computational Linguistics",
    url = "https://aclanthology.org/2025.acl-long.770/",
    doi = "10.18653/v1/2025.acl-long.770",
    pages = "15839--15853",
    ISBN = "979-8-89176-251-0"
}

@inproceedings{krishna-etal-2025-fact,
    title = "Fact, Fetch, and Reason: A Unified Evaluation of Retrieval-Augmented Generation",
    author = "Krishna, Satyapriya  and
      Krishna, Kalpesh  and
      Mohananey, Anhad  and
      Schwarcz, Steven  and
      Stambler, Adam  and
      Upadhyay, Shyam  and
      Faruqui, Manaal",
    editor = "Chiruzzo, Luis  and
      Ritter, Alan  and
      Wang, Lu",
    booktitle = "Proceedings of the 2025 Conference of the Nations of the Americas Chapter of the Association for Computational Linguistics: Human Language Technologies (Volume 1: Long Papers)",
    month = apr,
    year = "2025",
    address = "Albuquerque, New Mexico",
    publisher = "Association for Computational Linguistics",
    url = "https://aclanthology.org/2025.naacl-long.243/",
    doi = "10.18653/v1/2025.naacl-long.243",
    pages = "4745--4759",
    ISBN = "979-8-89176-189-6"
}

@inproceedings{heiden2011txm,
    title = "The {TXM} Platform: Building Open-Source Textual Analysis Software Compatible with the {TEI} Encoding Scheme",
    author = "Heiden, Serge",
    editor = "Otoguro, Ryo  and
      Ishikawa, Kiyoshi  and
      Umemoto, Hiroshi  and
      Yoshimoto, Kei  and
      Harada, Yasunari",
    booktitle = "Proceedings of the 24th Pacific Asia Conference on Language, Information and Computation",
    month = nov,
    year = "2010",
    address = "Tohoku University, Sendai, Japan",
    publisher = "Institute of Digital Enhancement of Cognitive Processing, Waseda University",
    url = "https://aclanthology.org/Y10-1044/",
    pages = "389--398"
}

@article{dahl2024large,
    author = {Dahl, Matthew and Magesh, Varun and Suzgun, Mirac and Ho, Daniel E},
    title = {Large Legal Fictions: Profiling Legal Hallucinations in Large Language Models},
    journal = {Journal of Legal Analysis},
    volume = {16},
    number = {1},
    pages = {64-93},
    year = {2024},
    month = {01},
    issn = {2161-7201},
    doi = {10.1093/jla/laae003},
    url = {https://doi.org/10.1093/jla/laae003},
    eprint = {https://academic.oup.com/jla/article-pdf/16/1/64/58336922/laae003.pdf},
}

@inproceedings{Turpin,
    author = {Turpin, Miles and Michael, Julian and Perez, Ethan and Bowman, Samuel R.},
    title = {Language models don't always say what they think: unfaithful explanations in chain-of-thought prompting},
    year = {2023},
    publisher = {Curran Associates Inc.},
    address = {Red Hook, NY, USA},
    booktitle = {Proceedings of the 37th International Conference on Neural Information Processing Systems},
    articleno = {3275},
    numpages = {14},
    location = {New Orleans, LA, USA},
    series = {NIPS '23},
    url = "https://dl.acm.org/doi/10.5555/3666122.3669397"
}

@misc{rouillard2026llmsscoremedicaldiagnoses,
      title={Can LLMs Score Medical Diagnoses and Clinical Reasoning as well as Expert Panels?}, 
      author={Amy Rouillard and Sitwala Mundia and Linda Camara and Michael Cameron Gramanie and Ziyaad Dangor and Ismail Kalla and Shabir A. Madhi and Kajal Morar and Marlvin T. Ncube and Haroon Saloojee and Bruce A. Bassett},
      year={2026},
      eprint={2604.14892},
      archivePrefix={arXiv},
      primaryClass={cs.LG},
      url={https://arxiv.org/abs/2604.14892}, 
}

@inproceedings{herrewijnen-etal-2026-bert,
    title = "{BERT}, are you paying attention? Attention regularization with human-annotated rationales",
    author = "Herrewijnen, Elize  and
      Nguyen, Dong  and
      Bex, Floris  and
      Gatt, Albert",
    editor = "Demberg, Vera  and
      Inui, Kentaro  and
      Marquez, Llu{\'i}s",
    booktitle = "Proceedings of the 19th Conference of the {E}uropean Chapter of the {A}ssociation for {C}omputational {L}inguistics",
    year = "2026",
    address = "Rabat, Morocco",
    publisher = "Association for Computational Linguistics",
    url = "https://aclanthology.org/2026.eacl-long.31/",
    doi = "10.18653/v1/2026.eacl-long.31",
    pages = "720--751",
    ISBN = "979-8-89176-380-7"
}

@misc{robinson_contrastive_2020,
	title = {Contrastive {Learning} with {Hard} {Negative} {Samples}},
	copyright = {arXiv.org perpetual, non-exclusive license},
	url = {https://arxiv.org/abs/2010.04592},
	doi = {10.48550/ARXIV.2010.04592},
	urldate = {2026-05-25},
	publisher = {arXiv},
	author = {Robinson, Joshua and Chuang, Ching-Yao and Sra, Suvrit and Jegelka, Stefanie},
	year = {2020},
	note = {Version Number: 2},
}

@inproceedings{karpukhin_dense_2020,
	address = {Online},
	title = {Dense {Passage} {Retrieval} for {Open}-{Domain} {Question} {Answering}},
	url = {https://aclanthology.org/2020.emnlp-main.550},
	doi = {10.18653/v1/2020.emnlp-main.550},
	language = {en},
	urldate = {2026-05-25},
	booktitle = {Proceedings of the 2020 {Conference} on {Empirical} {Methods} in {Natural} {Language} {Processing} ({EMNLP})},
	publisher = {Association for Computational Linguistics},
	author = {Karpukhin, Vladimir and Oguz, Barlas and Min, Sewon and Lewis, Patrick and Wu, Ledell and Edunov, Sergey and Chen, Danqi and Yih, Wen-tau},
	year = {2020},
	pages = {6769--6781},
}

@misc{xiong_approximate_2020,
	title = {Approximate {Nearest} {Neighbor} {Negative} {Contrastive} {Learning} for {Dense} {Text} {Retrieval}},
	copyright = {arXiv.org perpetual, non-exclusive license},
	url = {https://arxiv.org/abs/2007.00808},
	doi = {10.48550/ARXIV.2007.00808},
	urldate = {2026-05-25},
	publisher = {arXiv},
	author = {Xiong, Lee and Xiong, Chenyan and Li, Ye and Tang, Kwok-Fung and Liu, Jialin and Bennett, Paul and Ahmed, Junaid and Overwijk, Arnold},
	year = {2020},
	note = {Version Number: 2},
}

@inproceedings{gao_simcse_2021,
	address = {Online and Punta Cana, Dominican Republic},
	title = {{SimCSE}: {Simple} {Contrastive} {Learning} of {Sentence} {Embeddings}},
	shorttitle = {{SimCSE}},
	url = {https://aclanthology.org/2021.emnlp-main.552},
	doi = {10.18653/v1/2021.emnlp-main.552},
	language = {en},
	urldate = {2026-05-25},
	booktitle = {Proceedings of the 2021 {Conference} on {Empirical} {Methods} in {Natural} {Language} {Processing}},
	publisher = {Association for Computational Linguistics},
	author = {Gao, Tianyu and Yao, Xingcheng and Chen, Danqi},
	year = {2021},
	pages = {6894--6910}
}

@inproceedings{cai_hard_2022,
	address = {Atlanta GA USA},
	title = {Hard {Negatives} or {False} {Negatives}: {Correcting} {Pooling} {Bias} in {Training} {Neural} {Ranking} {Models}},
	isbn = {978-1-4503-9236-5},
	shorttitle = {Hard {Negatives} or {False} {Negatives}},
	url = {https://dl.acm.org/doi/10.1145/3511808.3557343},
	doi = {10.1145/3511808.3557343},
	language = {en},
	urldate = {2026-05-25},
	booktitle = {Proceedings of the 31st {ACM} {International} {Conference} on {Information} \& {Knowledge} {Management}},
	publisher = {ACM},
	author = {Cai, Yinqiong and Guo, Jiafeng and Fan, Yixing and Ai, Qingyao and Zhang, Ruqing and Cheng, Xueqi},
	month = oct,
	year = {2022},
	pages = {118--127},
}

@inproceedings{zhou_simans_2022,
	address = {Abu Dhabi, UAE},
	title = {{SimANS}: {Simple} {Ambiguous} {Negatives} {Sampling} for {Dense} {Text} {Retrieval}},
	shorttitle = {{SimANS}},
	url = {https://aclanthology.org/2022.emnlp-industry.56},
	doi = {10.18653/v1/2022.emnlp-industry.56},
	language = {en},
	urldate = {2026-05-25},
	booktitle = {Proceedings of the 2022 {Conference} on {Empirical} {Methods} in {Natural} {Language} {Processing}: {Industry} {Track}},
	publisher = {Association for Computational Linguistics},
	author = {Zhou, Kun and Gong, Yeyun and Liu, Xiao and Zhao, Wayne Xin and Shen, Yelong and Dong, Anlei and Lu, Jingwen and Majumder, Rangan and Wen, Ji-rong and Duan, Nan},
	year = {2022},
	pages = {548--559},
}

@inproceedings{wischounig_negative_2026,
	address = {Rabat, Morocco},
	title = {Negative {Sampling} {Techniques} in {Dense} {Retrieval}: {A} {Survey}},
	shorttitle = {Negative {Sampling} {Techniques} in {Dense} {Retrieval}},
	url = {https://aclanthology.org/2026.findings-eacl.157},
	doi = {10.18653/v1/2026.findings-eacl.157},
	language = {en},
	urldate = {2026-05-25},
	booktitle = {Findings of the {Association} for {Computational} {Linguistics}: {EACL} 2026},
	publisher = {Association for Computational Linguistics},
	author = {Wischounig, Laurin and Abdallah, Abdelrahman and Jatowt, Adam},
	year = {2026},
	pages = {3003--3020},
}

\appendix
\section{Appendix}
\label{sec:appendix}

\subsection{Related Work}
\subsubsection{Interpretable RAG}
\label{sec:appendix_relatedwork}
\citet{li-etal-2025-iris} introduces IRIS, an interpretable retrieval-augmented framework for long-document classification that segments documents into chunks, retrieves task-relevant embeddings using learnable query vectors, and aggregates them via linear attention with computation invariant to input length. The model achieves competitive performance on standard benchmarks and substantially outperforms long-context baselines on clinical risk prediction tasks, while providing intrinsic interpretability by explicitly exposing the retrieved evidence and latent factors underlying its predictions. \citet{deng-etal-2024-learning} introduces a neural retrieval and re-ranking framework that substantially improves downstream performance by optimizing evidence selection, particularly in long-document and multi-source settings. Experimental results demonstrate consistent gains over strong baselines, highlighting the dominant role of retrieval quality in model decisions. While the paper provides qualitative inspections of top-ranked evidence to illustrate relevance, interpretability remains largely implicit and post-hoc, as the approach does not explicitly model why specific evidence is retrieved or how retrieved content contributes to the final prediction. \citet{saxena2025rankingfreerag} propose METEORA, a rationale-driven, rank-free RAG framework in which rationales guide an unsupervised Evidence Chunk Selection Engine (ECSE) with adaptive cutoff via elbow detection, alongside a Verifier LLM that filters poisoned or inconsistent content. Evaluated across six benchmark datasets, METEORA achieves substantially higher recall and precision in evidence selection while using approximately 50\% fewer evidence chunks than state-of-the-art re-ranking approaches, resulting in a 33.34\% absolute improvement in generation accuracy over the strongest baseline; moreover, under adversarial conditions, it improves poison detection F1 from 0.10 to 0.44, demonstrating strong robustness and interpretability. Complementarily, \citet{liang-etal-2020-alice} introduce ALICE, an active learning framework that leverages contrastive natural language explanations (NLEs) to enhance data efficiency and interpretability by selecting examples based on model uncertainty and disagreement between model and human explanations, aligning reasoning in a joint embedding space through explanation-based acquisition functions and contrastive losses. Experiments on tasks such as NLI and commonsense reasoning show competitive performance with 30-40\% fewer labeled instances and improved robustness over uncertainty-only baselines. Finally, \citet{jacovi-etal-2021-contrastive} propose a contrastive explanation framework that projects internal representations into a latent space isolating features distinguishing facts from foils, enabling fine-grained attribution via causal interventions such as token masking and amnesic probing; evaluations on MultiNLI and BIOS reveal increased transparency, effective model debugging, and the exposure of biases including gender profession correlations.

\subsubsection{Hard Negative Selection}
In contrastive learning, hard negative selection involves choosing negative examples that are highly similar to positive ones, forcing the model to learn finer distinctions and thereby improving its discriminative performance \citep{robinson_contrastive_2020}. This idea has become central in dense retrieval, where models are commonly trained with in-batch negatives and lexically retrieved hard negatives, such as passages retrieved by BM25, a standard term-matching retrieval method, that overlap with the query but do not contain the answer \citep{karpukhin_dense_2020}. Subsequent work has shown that more informative negatives can be mined globally from the corpus rather than only from the current mini-batch \citep{xiong_approximate_2020}, and that linguistic supervision can define harder semantic contrasts: for example, supervised SimCSE uses natural language inference datasets by treating entailment pairs as positives and contradiction pairs as hard negatives \citep{gao_simcse_2021}. However, harder negatives are not always better: in retrieval settings, top-ranked unlabeled documents may be false negatives rather than genuinely irrelevant examples \citep{cai_hard_2022}, motivating methods that avoid both trivially easy negatives and overly hard negatives that may actually be relevant \citep{zhou_simans_2022}. Recent surveys of dense retrieval likewise emphasize that negative sampling strategy is a central design choice, spanning in-batch negatives, static lexical hard negatives, dynamic model-based mining, and synthetic negatives, while also highlighting the risk that highly ranked negatives may actually be unlabeled positives \citep{wischounig_negative_2026}.
CERA builds on this line of work by selecting hard negatives that are not merely topically similar but subjectively confusable with factual evidence, and by coupling contrastive retriever training with rationale-aligned attention so that the model learns to distinguish evidence-bearing tokens rather than merely coarse topical similarity.

\section{Hard Negative Selection Based on Subjectivity: A Triplets-Based Corpus Analysis}
\label{sec:ape_corpus_analysis}
\begin{figure*}[!htb]
    \centering
    \begin{minipage}[t]{0.32\linewidth}
        \centering
        \includegraphics[width=\linewidth]{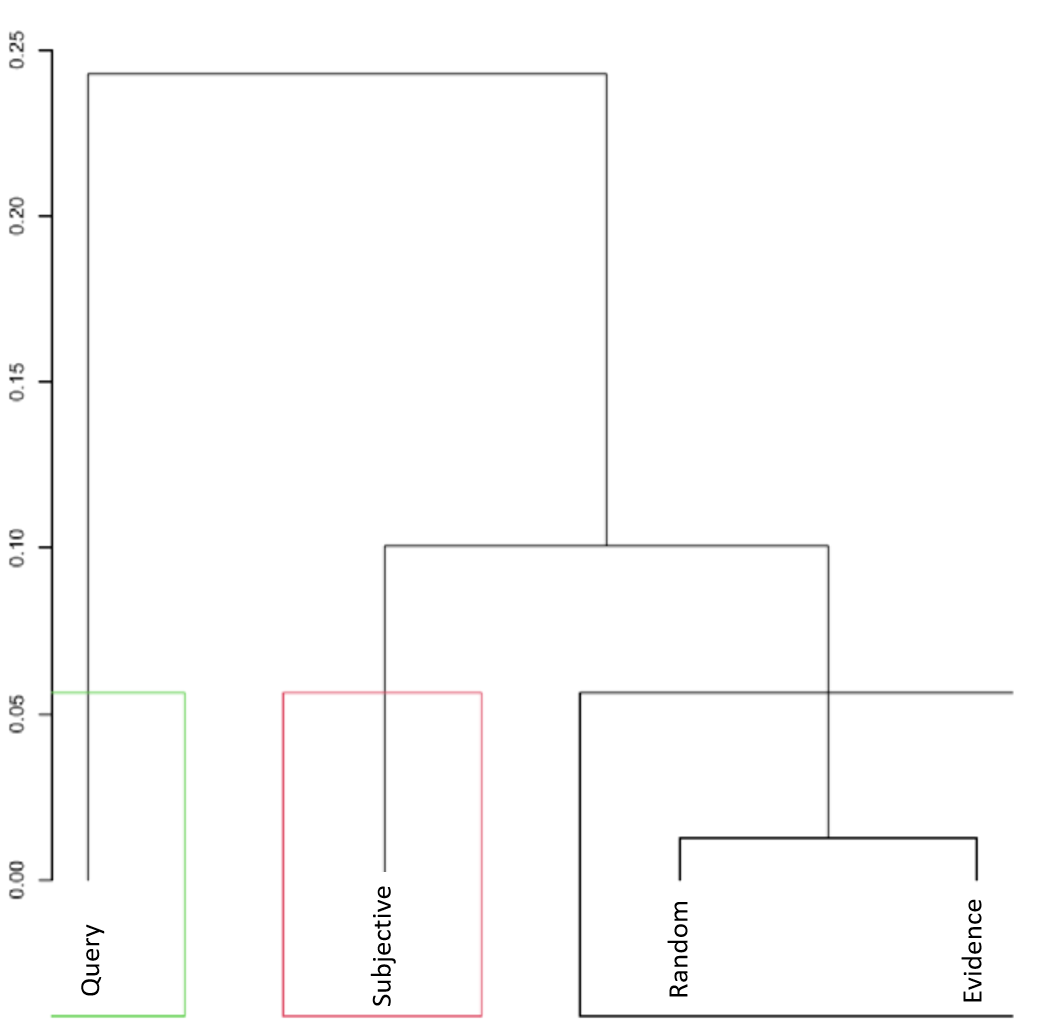}
    \end{minipage}
    \hfill
    \begin{minipage}[t]{0.67\linewidth}
        \centering
        \includegraphics[width=\linewidth]{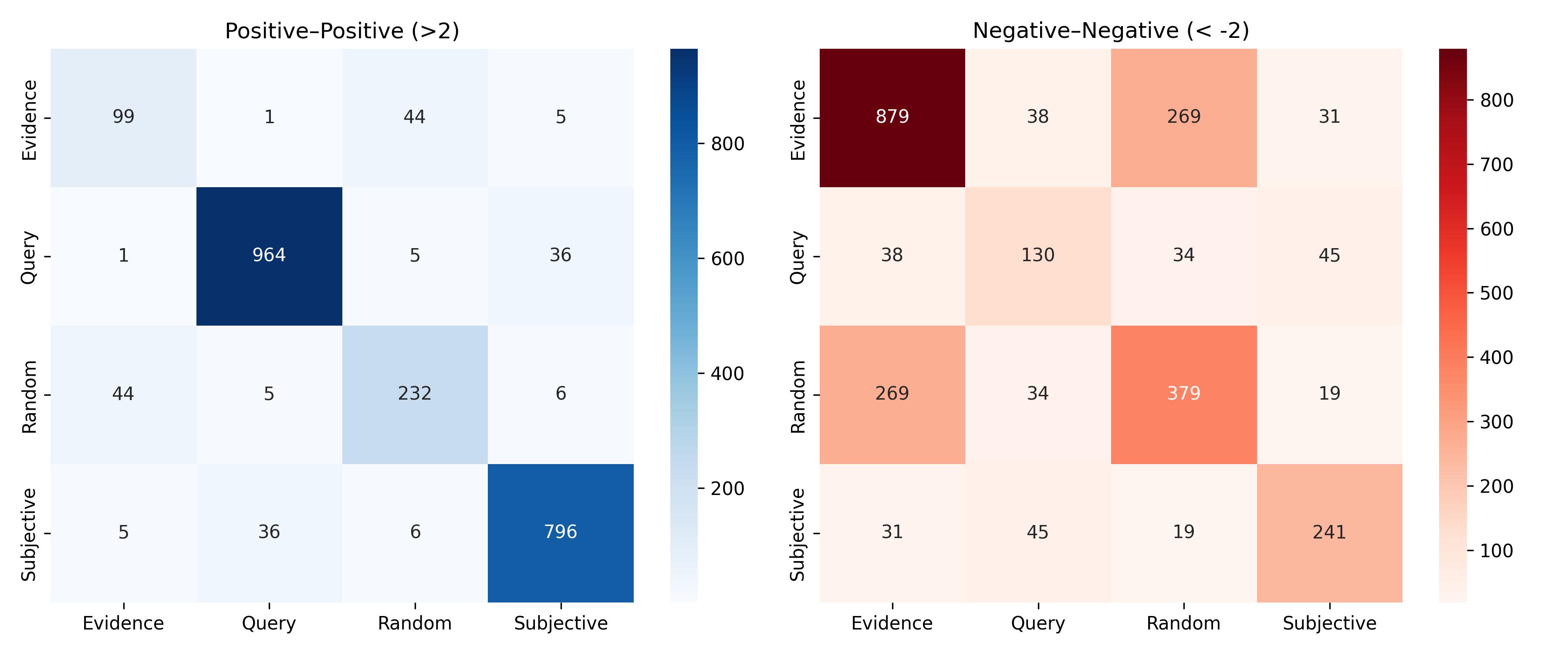}
    \end{minipage}
    \caption{Hierarchical clustering based on lemmas (left) and heatmaps showing pairwise co-occurrence of lemmas with significant specificity scores across text classes (right). Positive scores ($>2$), indicating over-representation, are shown on the right, while negative scores ($<-2$), indicating under-representation, are shown on the left.}
    \label{fig:lexical_analysis}
\end{figure*}

In this paper, we address limitations of existing hard negative selection methods, which often fail to capture complex semantic distinctions such as the difference between factual and non-factual content. To mitigate this issue, we propose a subjectivity-based hard negative selection strategy, where negatives are automatically identified from passages containing subjective or misleading explanations rather than verifiable evidence. This encourages the model to bring factual rationales closer in the embedding space while pushing subjective explanations apart, inducing representations aligned with evidential reasoning.

To explore the lexical patterns associated with the different text classes (i.e., \textit{query}, \textit{evidence}-positive evidence chunks, \textit{subjective}-negative subjective chunks, and \textit{random}-other chunks from the document), we used the TXM software \citep{heiden2011txm}, an open-source platform for textometric analysis that integrates statistical and linguistic tools for corpus exploration. We applied the TXM cluster analysis module through the Partition by Class feature, grouping sentences according to their assigned class. Clustering was performed on lemmas (canonical word forms) to capture general lexical tendencies beyond surface variation. 

The analysis used hierarchical agglomerative clustering based on a \textbf{lexical distance matrix} to assess \textbf{dissimilarities between positive and negative passages}. The resulting dendrogram (Figure~\ref{fig:lexical_analysis}, left) shows that the \textit{query} is the most lexically distinct, reflecting its comparative nature and independent source, while \textit{random} and \textit{positive} (evidence) display the highest lexical similarity. Lemma specificity analysis was conducted using TXM, which computes specificity scores via a hypergeometric model. Statistically significant scores (|score| > 2) were used to build pairwise co-occurrence matrices, visualized as heatmaps (Figure~\ref{fig:lexical_analysis}, right), capturing lexical overlap and divergence across classes. Results indicate that \textit{positive} (evidence contains fewer specific lemmas and largely overlaps with random, whereas query and subjective show more highly specific terms, indicating greater lexical diversity. Both query and evidence are characterized by statistical and comparative terminology, with higher specificity in evidence, which also exhibits greater use of numerals and specialized terms. The \textit{query} is marked by query-specific vocabulary, while \textit{negatives} (subjective) display greater grammatical variety, including auxiliary verbs and function words. 

For example:
\begin{itemize}
    \item \textit{query}: compare the effect of tomox versus folfox4 on hepatic disorders and asthenia.
    \item \textit{positive}: the two groups were significantly different in the hair shedding counts following hair washing (f (1,58) = 4 51, p = 0 038) due to a significant decrease among the mps-treated subjects (mps: d0: 27 13 ± 27; d90:16 47 ± 14; placebo: d0:23 4 ± 25, d90:21 87 ± 21)
    \item \textit{negative}: recruitment sites included community-based organizations and selected programs that provided services to the target population, such as the welfare
\end{itemize}

These examples illustrate the observed lexical differences, with technical terms being more specific to the query and positive texts. More specifically, numbers, comparison terms, and adverbs such as \textit{significantly} appear more frequently in the positive chunks. On the other hand, negative chunks tend to use a more general lexicon (i.e., less terminological) and focus more on methodological descriptions, without numerical results or statistical terms such as \textit{p}.

This corpus analysis shows that positive evidence chunks differ considerably in terms of lexicon from negative subjective chunks, being closer to random chunks from the texts. Query chunks are the most specific with respect to lemmas; however, Figure~\ref{fig:lexical_analysis} shows a slightly higher lexical intersection between subjective and query chunks when compared to the other two classes. This may be due to the more diverse vocabulary of the subjective class, whereas evidence chunks exhibit lower lexical similarity; when the same lexical items are used, they tend to be more consistent across chunks (e.g., statistical and terminological terms).

 
 As shown in Figure \ref{fig:lexical_analysis}, the fine-tuning strategy based on a subjectivity-aware hard negative selection criterion leads to a substantially improved embedding geometry. Compared to the self-learning setup, the fine-tuned Contriever model exhibits higher intra-class distances for both positives (Intra-Pos: 0.7766 vs. 0.5716) and negatives (Intra-Neg: 0.8141 vs. 0.5977), indicating a richer and more expressive representation space. More importantly, the increased inter-class distance (Inter: 0.8110 vs. 0.5901) demonstrates a clearer separation between positive and negative instances. This suggests that subjectivity-aware hard negative selection, explicitly contrasted with evidence-based positives, improves evidence alignment by reducing representational overlap and structuring embeddings for more accurate evidence-based retrieval.

\section{Ablation Studies}
\label{sec:appendix_ablation}

\subsection{Sensitivity Analysis of the Alignment Loss Weight $\lambda$}
To analyze the effect of the alignment loss weight, we perform an ablation study varying $\lambda$ in CERA$_{\text{alignment}}$. We compare $\lambda=0.01$ and $\lambda=0.05$ under both retrieval and explanation quality settings. Table~\ref{tab:cera_alignment_variants_app} reports ranking-based metrics, while Table~\ref{tab:lambda_ablation_app} shows plausibility and faithfulness results. These results highlight the impact of the alignment strength on both retrieval performance and rationale quality.

\begin{table}[!htb]
\centering
\scalefont{0.72}
\begin{tabular}{p{2.2cm}p{1.8cm}p{1.8cm}}
\hline
\textbf{Metrics} & 
$\mathbf{CERA_{\text{alignment}}^{\lambda=0.01}}$ & 
$\mathbf{CERA_{\text{alignment}}^{\lambda=0.05}}$ \\
\hline

\multicolumn{3}{l}{\textbf{Recall@K}} \\[2pt]
Recall@1  & \textbf{0.1898} & 0.1747 \\
Recall@3  & \textbf{0.4026} & 0.3703 \\
Recall@5  & \textbf{0.5391} & 0.5085 \\
Recall@10 & \textbf{0.7345} & 0.7110 \\
Recall@20 & \textbf{0.8880} & 0.8816 \\
Recall@50 & \textbf{0.9779} & 0.9789 \\

\hline
\multicolumn{3}{l}{\textbf{Precision@K}} \\[2pt]
Precision@1  & \textbf{0.2473} & 0.2291 \\
Precision@3  & \textbf{0.1795} & 0.1651 \\
Precision@5  & \textbf{0.1449} & 0.1369 \\
Precision@10 & \textbf{0.0995} & 0.0966 \\
Precision@20 & \textbf{0.0603} & 0.0599 \\
Precision@50 & \textbf{0.0265} & 0.0266 \\

\hline
\multicolumn{3}{l}{\textbf{NDCG@K}} \\[2pt]
NDCG@1  & \textbf{0.2473} & 0.2291 \\
NDCG@3  & \textbf{0.3357} & 0.3081 \\
NDCG@5  & \textbf{0.3962} & 0.3693 \\
NDCG@10 & \textbf{0.4647} & 0.4402 \\
NDCG@20 & \textbf{0.5073} & 0.4875 \\
NDCG@50 & \textbf{0.5270} & 0.5087 \\

\hline
\multicolumn{3}{l}{\textbf{MAP@K}} \\[2pt]
MAP@1  & \textbf{0.2473} & 0.2291 \\
MAP@3  & \textbf{0.2926} & 0.2668 \\
MAP@5  & \textbf{0.3287} & 0.3035 \\
MAP@10 & \textbf{0.3611} & 0.3367 \\
MAP@20 & \textbf{0.3752} & 0.3522 \\
MAP@50 & \textbf{0.3792} & 0.3565 \\

\hline
\textbf{MRR} & \textbf{0.41664} & 0.3934 \\
\hline
\end{tabular}
\caption{Ablation studies considering weighted CERA$_{\text{alignment}}$ variants with different $\lambda$ values.}
\label{tab:cera_alignment_variants_app}
\end{table}

\begin{table*}[!htb]
\centering
\footnotesize
\scalefont{0.80}
\setlength{\tabcolsep}{4pt}
\renewcommand{\arraystretch}{1.1}
\begin{tabular}{lcccccc}
\hline
& \multicolumn{4}{c}{Plausibility} & \multicolumn{2}{c}{Faithfulness} \\
\cline{2-5} \cline{6-7}
Model & IOU-F1 $\uparrow$ & Token P $\uparrow$ & Token R $\uparrow$ & Token F1 $\uparrow$ & Comp. $\uparrow$ & Suff. $\downarrow$ \\
\hline
CERA$_{\text{alignment}}$ $\lambda=0.01$ & 0.1570 & 0.7637 & 0.4131 & \textbf{0.5286} & \textbf{0.1173} & 0.1250 \\
CERA$_{\text{alignment}}$ $\lambda=0.05$ & \textbf{0.1701} & \textbf{0.7568} & \textbf{0.4163} & 0.5283 & 0.1113 & \textbf{0.0939} \\
\hline
\end{tabular}
\caption{Evaluation results comparing alignment variants of CERA$_{\text{alignment}}$ with different $\lambda$ values.}
\label{tab:lambda_ablation_app}
\end{table*}

As shown in Tables \ref{tab:cera_alignment_variants_app} and \ref{tab:lambda_ablation_app}, the ablation results reveal a consistent trade-off between retrieval effectiveness and alignment supervision strength. Increasing the alignment weight from $\lambda=0.01$ to $\lambda=0.05$ leads to systematic performance degradation across nearly all ranking metrics, including Recall@10, NDCG@10, MAP@10, and MRR. This behavior suggests that the alignment objective introduces a meaningful optimization bias in the representation space. A moderate alignment weight appears to act as a beneficial regularizer, encouraging the model to focus on semantically meaningful evidence while preserving retrieval quality. In contrast, stronger alignment supervision may over-constrain the encoder by forcing attention distributions to follow human factuality-based rationales too rigidly, thus reducing the model's flexibility to learn highly discriminative ranking representations.

Furthermore, the degradation is concentrated primarily in early-ranking metrics such as Recall@1, Precision@1, NDCG@5, MAP@10, and MRR, while the differences at larger retrieval cutoffs remain marginal (e.g., Recall@50). This suggests that stronger alignment supervision does not substantially impair the model's overall ability to retrieve relevant candidates, but mainly affects fine-grained ranking quality among top-ranked results. From a representation learning perspective, these findings indicate that alignment supervision preserves global semantic retrieval capacity while slightly reducing ranking sharpness at higher confidence positions.

Overall, the results demonstrate that the proposed approach for interpretable attention alignment through evidential inductive bias is not optimization-neutral: increasing alignment pressure influences retrieval behavior in measurable ways. The findings further suggest that moderate alignment strength provides a more favorable balance between retrieval effectiveness and rationale-guided supervision, highlighting the importance of carefully controlling the contribution of the alignment objective during training.

\subsection{Training Configuration and Hyperparameter Sensitivity Analysis}
Table \ref{tab:cera_vs_hardneg_app} presents an ablation study evaluating different training configurations of the proposed  CERA$_{\text{alignment}}$ model in comparison with a baseline based on hard negative selection. The experiments analyze the impact of varying batch sizes (4, 8, 16, and 32), learning rates ($1e^{-6}$ and $2e^{-6}$), the use of optimizers, and the number of training epochs.

The results in Table~\ref{tab:cera_vs_hardneg_app} reveal a consistent and substantial advantage of  CERA$_{\text{alignment}}$ over the HardNeg$_{\text{base}}$, baseline across all evaluated training configurations and ranking metrics. Regardless of the \textit{batch size}, \textit{learning rate}, optimization (scheduler)\footnote{\url{https://huggingface.co/transformers/v4.2.2/main_classes/optimizer_schedules.html}}, or training epoch, the proposed CERA$_{\text{alignment}}$ systematically improves retrieval effectiveness. The gains are particularly pronounced in Recall@10, where CERA$_{\text{alignment}}$ achieves improvements of approximately 5--7 absolute points over the hard negative baseline, indicating a significantly stronger ability to retrieve relevant candidates within the top-ranked positions.

A similar trend can be observed for NDCG@10, MAP@10, and MRR, suggesting that the improvements are not limited to recall-oriented retrieval behavior, but also extend to ranking quality and early precision. Interestingly, the improvements remain stable across distinct optimization settings, which indicates that the proposed alignment strategy is robust to hyperparameter variations. This stability is especially important in retrieval-based NLP systems, where performance often fluctuates considerably under different batch sizes or learning rate schedules. In contrast, the HardNeg$_{\text{base}}$, results exhibit relatively narrow variance and appear to plateau earlier, suggesting limited benefits from additional optimization tuning.

The strongest overall configuration for CERA$_{\text{alignment}}$ is obtained with \textit{batch8\_1e-6\_sched}, which achieves the highest Recall@10 (0.70424) and NDCG@10 (0.42823), while \textit{batch16\_1e-6\_sched} produces the best MAP@10 (0.32535) and MRR (0.38419). This pattern suggests a trade-off between broader retrieval coverage and ranking refinement. Smaller or medium batch sizes appear to favor recall-oriented behavior, whereas larger batches combined with scheduling slightly improve ranking consistency and top-position relevance. Nevertheless, even configurations that are not individually optimal still outperform the hard negative baseline by a considerable margin, reinforcing that the primary source of improvement stems from the proposed subjectivity-aware alignment mechanism rather than from hyperparameter tuning alone.

Overall, these findings suggest that subjectivity-based negatives provide a more informative contrastive signal than conventional hard negatives. While hard negative mining typically focuses on lexical or semantic proximity, the proposed approach appears to better capture fine-grained alignment properties relevant to the downstream retrieval task. Consequently, the model learns more discriminative representations, leading to consistent gains in both retrieval coverage and ranking quality across all evaluated configurations.

\section{LLM Evaluation Prompt}
\label{sec:appendix_prompt}
To assess the quality of a span, we use the prompt shown in Figure \ref{fig:judge-prompt}, providing the LLM with the task, a detailed scoring system, the retrieved span at rank $R$, as well as the actual true positives (gold spans) as ground truth references. We deliberately did not choose few shot examples in the prompt, as these can bias LLMs to prefer certain answers over others \cite{Turpin}. To increase reproducibility, we inferred from all LLMs with a temperature set to 0. We intentionally selected models from multiple providers to ensure a diversity of perspectives among the judges. 

While our LLM-based evaluation is inherently limited by the models’ internal knowledge and contextual reasoning, prior work \cite{rouillard2026llmsscoremedicaldiagnoses} suggests that these systems can still provide meaningful assessment signals. In particular, \citet{rouillard2026llmsscoremedicaldiagnoses} show that LLM juries can act as reliable proxies for expert diagnostic evaluation in medical settings, highlighting their potential as evaluative tools.

\begin{figure*}[t]
\centering
\begin{tcolorbox}[
    enhanced,
    colback=gray!5,
    colframe=black!70,
    boxrule=0.5pt,
    arc=2pt,
    left=6pt,
    right=6pt,
    top=6pt,
    bottom=6pt,
    fonttitle=\bfseries,
    title=LLM-as-a-Judge Evaluation Prompt (Including System Prompt),
    coltitle=white,
    colbacktitle=black!70,
]
\small
\ttfamily
\raggedright
You are an expert assessor for biomedical information retrieval.

\medskip
Given a query, one or more gold-standard reference spans, and a retrieved span, evaluate the retrieved span based on BOTH:
\begin{enumerate}\itemsep0pt
    \item Factual consistency with the gold span(s) (does it contain correct information?)
    \item Usefulness for answering the query
\end{enumerate}
 
When multiple gold spans are provided, treat them collectively as the reference: the retrieved span should be judged against the union of key facts expressed across all gold spans.
 
\medskip
Scoring rules (STRICT):
\begin{itemize}\itemsep0pt
    \item 0 = Factually incorrect OR contradicts the gold span(s), even if relevant
    \item 1 = Factually consistent but mostly irrelevant or contains no useful information for the query
    \item 2 = Factually consistent and partially useful, but incomplete or missing key information from the gold span(s)
    \item 3 = Factually consistent and highly useful; captures the key facts from the gold span(s) needed to answer the query
\end{itemize}
 
\medskip
Important guidelines:
\begin{itemize}\itemsep0pt
    \item Factual correctness is more important than relevance
    \item Any contradiction or clearly wrong statement $\rightarrow$ score 0
    \item Do NOT reward spans that are only topically similar but contain no concrete facts
    \item Prefer spans that contain specific, correct, and relevant information
\end{itemize}
 
\medskip
Reply with ONLY a single digit: 0, 1, 2, or 3. No explanation.

\medskip
\#\# Query: \{\}

\medskip
\#\# Gold span: \{\}

\medskip
\#\# Retrieved span: \{\}
\end{tcolorbox}
\caption{Prompt used for LLM-as-a-judge evaluation of retrieved spans against gold-standard references. The first sentence is provided as a system prompt to the LLMs. "\{\}" are filled with the respective query, span, and gold span or gold spans (with adjusted prompt structure) if multiple exist for a given test instance.}
\label{fig:judge-prompt}
\end{figure*}

\end{document}